%% file: main.tex
\documentclass[11pt,letterpaper]{mystyle}

\usepackage[all]{hypcap}
\usepackage[svgnames]{xcolor}
\usepackage[comma,authoryear,compress]{natbib}
\usepackage{hyperref}[citecolor=lightblue]

\hypersetup{
    colorlinks = true,
    citecolor = {YaleBlue},
}

\usepackage[utf8]{inputenc} 
\usepackage[T1]{fontenc}    
\usepackage{booktabs}       
\usepackage{amsfonts}       
\usepackage{nicefrac}       
\usepackage{microtype}      
\usepackage{xcolor}         
\definecolor{VeriGateSOne}{HTML}{4A8330}
\definecolor{VeriGateSTwo}{HTML}{C94A4A}
\definecolor{VeriGateSThree}{HTML}{3A71D0}
\colorlet{good}{VeriGateSOne}
\colorlet{bad}{VeriGateSTwo}
\colorlet{accent}{VeriGateSThree}
\DeclareRobustCommand{\verigatebadge}[2]{%
  \kern0.03em\tikz[baseline=(badge.base)]\node[%
    circle,%
    draw=none,%
    fill=#1,%
    text=black,%
    font=\bfseries\tiny,%
    minimum size=1.42em,%
    inner sep=0pt%
  ] (badge) {#2};\kern0.02em%
}
\DeclareRobustCommand{\badge}[2]{\verigatebadge{#1}{#2}}
\usepackage{tcolorbox}

\usepackage{xspace}
\newcommand{\ours}{\textsc{PRISM}\xspace}
\newcommand{\oursfull}{Precision Ranking for Improved Step Modeling\xspace}

\usepackage{amsmath}
\usepackage{amssymb}
\usepackage{mathtools}
\usepackage{amsthm}
\usepackage{makecell}
\usepackage{multirow}
\usepackage{enumitem}
\usepackage{tikz}
\usepackage{booktabs}
\usepackage{wrapfig}
\usetikzlibrary{arrows.meta, positioning, shapes.geometric, fit}
\usepackage{circledsteps}
\usepackage{subcaption}
\usepackage{algorithm}
\usepackage{algorithmicx}
\usepackage{algpseudocode}
\usepackage{microtype}
\usepackage{graphicx}
\expandafter\def\csname ver@subfig.sty\endcsname{}
\usepackage{booktabs} %
\usepackage{float}
\usepackage{bigstrut}

\usepackage{mathrsfs}
\usepackage{nicefrac}
\usepackage{dsfont}
\usepackage{enumitem}
\usepackage{subcaption}
\usepackage{graphicx,subfig}
\usepackage{cleveref}
\usepackage{bxcoloremoji}

\usepackage{float}

\usepackage[utf8]{inputenc} %
\usepackage[T1]{fontenc}    %
\usepackage{hyperref}       %
\usepackage{url}            %
\usepackage{booktabs}       %
\usepackage{amsfonts}       %
\usepackage{nicefrac}       %
\usepackage{microtype}      %
\usepackage{graphicx}
\usepackage{subcaption} 

\usepackage{amssymb}
\usepackage{fdsymbol}
\usepackage{wrapfig}
\usepackage{lipsum}
\usepackage{enumitem}
\usepackage{stackengine}
\usepackage[font=small,labelfont=bf]{caption}
\usepackage{color}
\usepackage{adjustbox}

\usepackage{rotating}
\usepackage{makecell}

\input{macro.sty}

\newtcolorbox{AIbox}[2][]{aibox,title=#2,#1}
\definecolor{lightblue}{rgb}{0.22,0.45,0.70}%
\definecolor{Gray}{gray}{0.95}
\definecolor{Cornsilk}{rgb}{1.0, 0.97, 0.86}

\usepackage{amsmath}

\usepackage[all]{hypcap}

\title{The Hidden Bias of Process Reward Models:\\ \ours for Rewarding the Right Reasoning}

\runningtitle{The Hidden Bias of Process Reward Models: \ours for Rewarding the Right Reasoning}

\author{
  Aakriti Agrawal$^{1}$,
  Souradip Chakraborty$^{1}$, 
  Armin Saghafian$^{1}$,
  Nihal Sharma$^{2}$, 
  Rizal Fathony$^{2}$, 
  Nam H Nguyen$^{2}$, 
  C. Bayan Bruss$^{2}$, 
  Amrit Singh Bedi$^{3}$, and
  Furong Huang
}

\affil[1]{University of Maryland, College Park}
\affil[2]{Capital One}
\affil[3]{University of Central Florida}

\correspondingauthor{Aakriti Agrawal; Email \href{mailto:agrawal5@umd.edu}{agrawal5@umd.edu}}

\begin{document}
\hbadness=10000
\hfuzz=1pt

\begin{abstract}
\input{sections/0.abstract}
\vspace{3mm}

\github{} \textbf{Code Repository}: \href{https://github.com/Aakriti05/PRISM.git}{https://github.com/Aakriti05/PRISM}

\coloremojicode{1F4E7} \textbf{Contact}: \href{mailto:agrawal5@umd.edu}{agrawal5@umd.edu}
\end{abstract}

\maketitle

\input{sections/1.Introduction}
\input{sections/2.Problem}

\input{sections/3.Method}
\input{sections/4.Results_Experiments}
\input{sections/5.RelatedWorks}
\input{sections/6.Conclusion_Limitation}

\section{Acknowledgement}
Agrawal, Chakraborty and Huang are supported by DARPA HR001124S0029-AIQ-FP-019,  National Science Foundation TRAILS Institute (2229885). Private support was provided by Open Philanthropy and Apple. The Authors acknowledge the National Artificial Intelligence Research Resource (NAIRR) Pilot and [insert the resources supporting your project here] for contributing to this research result.

\newpage
\bibliography{references}

\input{sections/Appendix}
\end{document}

%% file: sections/0.abstract.tex
Process Reward Models (PRMs) improve credit assignment for reasoning by providing step-level feedback. However, we identify a hidden bias in PRMs caused by severe imbalance in step-level training data. Standard cross-entropy training amplifies this bias, causing PRMs to overcredit plausible but incorrect steps and produce high false-positive rates. We show that these false positives have an asymmetric downstream effect: false negatives mainly slow exploration, whereas false positives actively steer Best-of-$N$ selection, guided decoding, and policy optimization toward flawed reasoning. This suggests that PRM training should shift from pointwise label fitting to reliable relative comparisons. To address this, we propose \ours~(\oursfull), a policy-aware PRM training framework that learns from contrastive step-level comparisons and hard negatives generated by a temporal lookahead strategy, requiring no new human labels. We further use a difficulty-aware curriculum to optimize the contrastive step margin. Across PRMBench~\citep{song2025prmbench} and ProcessBench~\citep{zheng2025processbench}, \ours substantially reduces false positives ($\downarrow$ 22\% on PRMBench) and improves macro F1 over strong discriminative PRMs. When applied to policy optimization and search tasks, including guided decoding and Best-of-$N$ selection, it consistently improves accuracy (up to 22\% for guided decoding and 33\% for Best-of-$N$) and robustness. More broadly, trustworthy process supervision is not just about assigning high rewards, but about rewarding the right reasoning for the right reasons.

%% file: sections/1.Introduction.tex
\providecommand{\badge}[2]{\textbf{(#2)}}
\providecommand{\PRMBench}{PRMBench}
\providecommand{\ProcessBench}{ProcessBench}

\begin{figure}[!htbp]
    \centering
    \resizebox{\textwidth}{!}{\input{Figures/teaser_aak.tex}}
    \vspace{-0.8em}
    \caption{Baseline PRMs can overcredit plausible but incorrect steps, a form of hidden PRM bias that downstream search and policy learning can exploit. \ours~\badge{good}{1} converts existing step annotations into matched comparisons, \badge{bad}{2} augments them with temporal-lookahead hard negatives using future steps $s(>t)$ (hard negatives are more confusing but helps in learning better reasoning), and \badge{accent}{3} trains with a Step-Contrastive objective plus a simple curriculum. The result is substantially lower FPR on \PRMBench ($69.3\%\rightarrow47.1\%$) and more reliable guided search, Best-of-$N$ selection, and policy optimization.}
    \label{fig:intro}
    \vspace{-1.2em}
\end{figure}
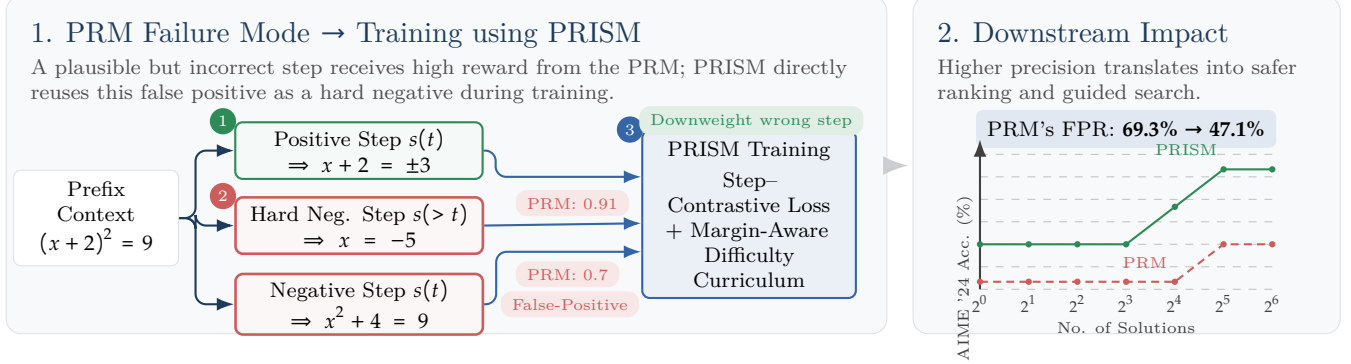

\section{Introduction}\label{sec:intro}
\vspace{-2mm}

Process Reward Models (PRMs) \citep{wang2023math, zhang2025lessons} have become a key ingredient in modern reasoning systems \citep{guo2025deepseek, shao2024deepseekmath, jaech2024openai} because they score intermediate reasoning steps rather than only final answers. This denser supervision improves credit assignment in long-horizon tasks \citep{lambert2024tulu} and is especially valuable in mathematics \citep{lightman2023let, chen2024alphamath, zhang2024rest}, program synthesis, and planning, where final success depends on many locally correct decisions. Yet current PRMs exhibit a hidden and consequential failure mode: across both discriminative and generative formulations \citep{zhang2025lessons, wang2023math, khalifa2025process, zhao2025genprm}, they often assign high scores to plausible but incorrect steps, producing elevated false-positive (FP) rates (Table~\ref{tab:intro_SOTA_PRMs}). This failure mode is referred to here as \emph{overcredit bias}.

\textbf{Sources of overcredit bias.} Overcredit bias is partly rooted in the structure of process-supervision data. Even incorrect steps frequently contain many locally correct steps before the first decisive error, so correct step labels can outnumber incorrect ones even within ultimately wrong trajectories (e.g., PRM800k; Figure~\ref{fig:data_imbalance})~\citep{lightman2023let}. This skews the step-level training distribution toward positives. When PRMs are then trained with standard \emph{pointwise} objectives such as cross-entropy on imbalanced and noisy step annotations, they tend to inherit the same bias, rewarding fluency and local plausibility more than actual correctness. The result is exactly the error pattern that matters most downstream: incorrect steps that receive undeservedly high scores.

\textbf{Why false positives matter most.} This failure mode is especially harmful because PRMs are not merely offline evaluators; they are used inside downstream policy learning and inference-time search. A false negative may discard a correct step and force the system to explore more. A false positive is worse: it can actively steer the policy onto an incorrect branch. In guided decoding, the model may keep expanding a flawed partial solution; in Best-of-$N$ selection, it may choose a wrong but convincing trajectory; and in reinforcement learning, it may reinforce spurious reasoning patterns. False positives therefore do not remain local prediction errors---they become a direct mechanism for bias-induced policy misalignment (Figure~\ref{fig:intro}).

\vspace{-1mm}

\textbf{A policy-aware view of PRM quality.} These observations suggest that PRMs should be evaluated and trained through the lens of downstream use. The goal is not simply to maximize average step-level accuracy, but to learn scores that reliably rank correct continuations above incorrect yet plausible ones. From this perspective, false positives and false negatives are not symmetric: false negatives primarily reduce recall and slow search, whereas false positives can make incorrect reasoning appear rewarding and thereby mislead the policy itself. This leads to a central question: \textit{can false positives be reduced without sacrificing the recall needed for exploration and search?} Answering it requires solving three practical challenges: \textbf{(C1)} learning an objective that optimizes relative ranking rather than pointwise label fit, \textbf{(C2)} constructing informative positive--negative comparisons from imbalanced step annotations, and \textbf{(C3)} stabilizing training when the most informative negatives are also the hardest.

\textbf{The proposed method: \ours.} \ours (\oursfull) is a label-efficient, architecture-agnostic training recipe that addresses these challenges without new human annotations. First, instead of pointwise classification, PRMs are trained with a \emph{Step-Contrastive} objective that optimizes the relative score of correct versus incorrect steps. Second, \emph{matched} positive--negative pairs are constructed from existing step annotations, and hard negatives are generated via \emph{temporal lookahead}, treating future steps as negatives for earlier contexts to expose overcredit errors. Third, optimization is stabilized with a simple difficulty-aware curriculum. Together, these ingredients shift PRM learning from pointwise label fitting to calibrated \emph{preference margins}, directly targeting the false-positive overcredit that undermines downstream alignment.

\textbf{Empirical validation.} On PRMBench \citep{song2025prmbench} and ProcessBench \citep{zheng2025processbench}, \ours consistently lowers FPR and improves macro F1 over strong baselines. More importantly, when deployed in guided decoding, Best-of-$N$ selection, and policy optimization tasks, the resulting PRMs deliver higher downstream accuracy and robustness. These gains narrow the gap between offline reward-model metrics and the behavior of the policies that use them.

\textbf{Contributions.} \textbf{(1) Precision-first analysis.} The analysis formalizes how step-level overcredit compounds across trajectories and proves a sharp asymmetry for Best-of-$N$: the false-positive rate determines an asymptotic alignment ceiling through precision, whereas false negatives primarily increase sample complexity. \textbf{(2) A comparison-based training recipe.} \ours provides a label-efficient, architecture-agnostic method that converts existing step labels into matched comparisons, trains with a Step-Contrastive objective, and combines temporal-lookahead hard negatives with a simple curriculum. \textbf{(3) Downstream-aligned gains.} Results on PRMBench \citep{song2025prmbench}, ProcessBench \citep{zheng2025processbench}, guided decoding, Best-of-$N$ selection, and policy training show that reducing overcredit yields more reliable PRMs and stronger policy alignment without new human labels.

Overall, these findings argue for \emph{precision-first} process supervision: trustworthy reasoning systems require not just dense rewards, but rewards that distinguish genuinely correct reasoning from merely plausible traces.

%% file: Figures/teaser_aak.tex
\begin{tikzpicture}[x=1cm,y=1cm, >=Latex, font=\sffamily]
    \definecolor{primary}{RGB}{31,59,91}        
    \definecolor{accent}{RGB}{52,101,164}       
    \definecolor{good}{RGB}{46,139,87}          
    \definecolor{bad}{RGB}{205,92,92}           
    \definecolor{bglight}{RGB}{248,249,250}     
    \definecolor{boxborder}{RGB}{222,226,230}   
    \definecolor{subtext}{RGB}{80,80,80}        

    \tikzstyle{panel} = [
        draw=boxborder,
        fill=bglight,
        rounded corners=2mm,
        inner sep=0pt
    ]
    \tikzstyle{header} = [
        font=\bfseries,
        text=primary,
        anchor=north west
    ]
    \tikzstyle{subtext} = [
        font=\scriptsize,
        text=subtext,
        align=left,
        anchor=north west
    ]
    \tikzstyle{nodebox} = [
        draw=boxborder,
        fill=white,
        rounded corners=1mm,
        align=center,
        font=\scriptsize,
        inner sep=1.5mm
    ]
    \tikzstyle{pill} = [
        rounded corners=1.5mm,
        inner xsep=1.5mm,
        inner ysep=1mm,
        font=\tiny\bfseries,
        align=center
    ]

    \node[
        fill=primary,
        text=white,
        rounded corners=1.5mm,
        minimum width=17.8cm,
        minimum height=0.7cm,
        font=\bfseries
    ] at (8.9, 5.3)
    {PRISM Reduces False--Positives In Process Rewards Models For Better Policy Learning.};

    \node[
        panel,
        minimum width=11.75cm,
        minimum height=4.5cm,
        anchor=south west
    ] (p12) at (0,0) {};

    \node[header] at (0.2,4.3)
    {1. PRM Failure Mode $\rightarrow$ Training using PRISM};

    \node[subtext, text width=11.1cm] at (0.2,3.8)
    {A plausible but incorrect step receives high reward from the PRM; PRISM directly reuses this false positive as a hard negative during training.};

    \node[
        nodebox,
        text width=1.9cm,
        anchor=west
    ] (ctx) at (0.10,1.55)
    {\textbf{Prefix Context}\\[0.5mm] $(x+2)^2 = 9$};

    \node[
        nodebox,
        draw=good,
        fill=good!5,
        line width=0.8pt,
        text width=3.0cm,
        inner ysep=0.8mm,
        anchor=west
    ] (pos) at (3.05,2.45)
    {
        \textbf{Positive Step} $s(t)$\\[0.2mm]
        $\Rightarrow x+2 = \pm 3$
    };

    \node[
        circle,
        fill=good,
        text=white,
        font=\tiny\bfseries,
        inner sep=1pt,
        minimum size=3.5mm
    ] at ([xshift=-1.5mm, yshift=0mm]pos.north west) {1};

    \node[
        nodebox,
        draw=bad,
        fill=bad!5,
        line width=1.0pt,
        text width=3.0cm,
        inner ysep=0.8mm,
        anchor=west
    ] (neg) at (3.05,1.45)
    {
        \textbf{Hard Neg. Step} $s(>t)$\\[0.2mm]
        $\Rightarrow x = -5$
    };

    \node[
        pill,
        fill=bad!15,
        text=bad,
        anchor=west
    ] at (6.80,1.75) {PRM: 0.91};

    \node[
        circle,
        fill=bad,
        text=white,
        font=\tiny\bfseries,
        inner sep=1pt,
        minimum size=3.5mm
    ] at ([xshift=-1.5mm, yshift=0mm]neg.north west) {2};

    \node[
        nodebox,
        draw=bad,
        fill=bad!5,
        line width=1.0pt,
        text width=3.0cm,
        inner ysep=0.8mm,
        anchor=west
    ] (negtwo) at (3.05,0.40)
    {
        \textbf{Negative Step} $s(t)$\\[0.2mm]
        $\Rightarrow x^2 + 4 = 9$
    };

    \node[
        pill,
        fill=bad!15,
        text=bad,
        anchor=west
    ] at (6.80,0.80) {PRM: 0.7};

    \node[
        pill,
        fill=bad!15,
        text=bad,
        anchor=west
    ] at (6.60,0.40) {False-Positive};

    \node[
        nodebox,
        draw=accent,
        fill=accent!10,
        line width=0.8pt,
        text width=2.55cm,
        minimum height=1.65cm,
        anchor=west
    ] (loss) at (8.48,1.60)
    {
        \textbf{PRISM Training}\\[0.7mm]
        Step--Contrastive Loss\\
        + Margin-Aware\\
        Difficulty Curriculum
    };

    \node[
        circle,
        fill=accent,
        text=white,
        font=\tiny\bfseries,
        inner sep=1pt,
        minimum size=3.5mm
    ] at ([xshift=-1.5mm, yshift=0mm]loss.north west) {3};

    \node[
        pill,
        fill=good!15,
        text=good,
        anchor=south
    ] at (9.95,2.65)
    {Downweight wrong step};

    \draw[->, primary, thick, rounded corners]
        (ctx.east) -- ++(0.22,0) |- (pos.west);

    \draw[->, primary, thick, rounded corners]
        (ctx.east) -- ++(0.22,0) |- (neg.west);

    \draw[->, primary, thick, rounded corners]
        (ctx.east) -- ++(0.22,0) |- (negtwo.west);

    \draw[->, accent, thick, rounded corners]
        (pos.east) -- ++(0.16,0) |- ([yshift=5mm]loss.west);

    \draw[->, accent, thick, rounded corners]
        (neg.east) -- ([yshift=-1.2mm]loss.west);

    \draw[->, accent, thick, rounded corners]
        (negtwo.east) -- ++(0.16,0) |- ([yshift=-5mm]loss.west);

    \node[
        panel,
        minimum width=5.7cm,
        minimum height=4.5cm,
        anchor=south west
    ] (p3) at (12.1,0) {};

    \node[header] at (12.3,4.3) {2. Downstream Impact};

    \node[subtext, text width=5.2cm] at (12.3,3.8)
    {Higher precision translates into safer ranking and guided search.};

    \node[
        nodebox,
        draw=none,
        fill=accent!15,
        anchor=north,
        font=\scriptsize
    ] at (14.95,3.0)
    {\textbf{PRM's FPR}: $\mathbf{69.3\% \rightarrow 47.1\%}$};

    \begin{scope}[shift={(13.0,0.6)}]
        \draw[lightgray, very thin, dashed] (0,0)   -- (3.9,0);
        \draw[lightgray, very thin, dashed] (0,0.3) -- (3.9,0.3);
        \draw[lightgray, very thin, dashed] (0,0.6) -- (3.9,0.6);
        \draw[lightgray, very thin, dashed] (0,0.9) -- (3.9,0.9);
        \draw[lightgray, very thin, dashed] (0,1.2) -- (3.9,1.2);
        \draw[lightgray, very thin, dashed] (0,1.5) -- (3.9,1.5);
        \draw[lightgray, very thin, dashed] (0,1.8) -- (3.9,1.8);

        \draw[->, darkgray, thick]
            (0,0) -- (0,2.0)
            node[left, font=\tiny, rotate=90, yshift=2mm, xshift=-6mm]
            {AIME '24 Acc. (\%)};

        \draw[bad, thick, densely dashed]
            (0,0.1) -- (0.65,0.1) -- (1.3,0.1) -- (1.95,0.1)
            -- (2.6,0.1) -- (3.25,0.6) -- (3.9,0.6);

        \fill[bad]
            (0,0.1) circle (1.2pt)
            (0.65,0.1) circle (1.2pt)
            (1.3,0.1) circle (1.2pt)
            (1.95,0.1) circle (1.2pt)
            (2.6,0.1) circle (1.2pt)
            (3.25,0.6) circle (1.2pt)
            (3.9,0.6) circle (1.2pt);

        \draw[good, thick]
            (0,0.6) -- (0.65,0.6) -- (1.3,0.6) -- (1.95,0.6)
            -- (2.6,1.1) -- (3.25,1.6) -- (3.9,1.6);

        \fill[good]
            (0,0.6) circle (1.2pt)
            (0.65,0.6) circle (1.2pt)
            (1.3,0.6) circle (1.2pt)
            (1.95,0.6) circle (1.2pt)
            (2.6,1.1) circle (1.2pt)
            (3.25,1.6) circle (1.2pt)
            (3.9,1.6) circle (1.2pt);

        \node[font=\tiny\bfseries, text=bad, anchor=south]
            at (2.2,0.15) {PRM};

        \node[font=\tiny\bfseries, text=good, anchor=south east]
            at (3.3,1.65) {PRISM};

        \node[font=\tiny, text=darkgray, anchor=north, inner sep=2pt]
            at (0,0) {$2^0$};
        \node[font=\tiny, text=darkgray, anchor=north, inner sep=2pt]
            at (0.65,0) {$2^1$};
        \node[font=\tiny, text=darkgray, anchor=north, inner sep=2pt]
            at (1.3,0) {$2^2$};
        \node[font=\tiny, text=darkgray, anchor=north, inner sep=2pt]
            at (1.95,0) {$2^3$};
        \node[font=\tiny, text=darkgray, anchor=north, inner sep=2pt]
            at (2.6,0) {$2^4$};
        \node[font=\tiny, text=darkgray, anchor=north, inner sep=2pt]
            at (3.25,0) {$2^5$};
        \node[font=\tiny, text=darkgray, anchor=north, inner sep=2pt]
            at (3.9,0) {$2^6$};

        \node[font=\tiny, text=darkgray, anchor=north]
            at (1.95,-0.3) {No. of Solutions};
    \end{scope}

    \draw[
        ->,
        line width=1.5pt,
        draw=boxborder!80!gray
    ] (11.8,2.25) -- (12.05,2.25);

\end{tikzpicture}

%% file: sections/2.Problem.tex
\vspace{-2mm}
\section{Problem Formulation: PRM Bias and Its Impact on Policy Learning}
\vspace{-2mm}

Let $x$ denote the prompt and $y=(y_1, y_2, \ldots, y_T)$ a reasoning chain with $T$ steps. At step $t$, the prefix $y_{<t}=(y_1, \ldots, y_{t-1})$ is the partial reasoning trace and $y_t$ is a candidate next step. A Process Reward Model (PRM) is a step-level verifier that gives scalar reward $r_{\theta}(x, y_{<t}, y_t)$, which measures how appropriate or correct the step $y_t$ is in the context $(x, y_{<t}, y_t)$. This step-level score is then used by downstream policy optimization or search procedures as discussed later.
\vspace{-2mm}
\subsection{Key Observation: Hidden bias in PRMs}
\vspace{-2mm}
\begin{wrapfigure}{r}{0.3\textwidth}
    \centering
    \vspace{-3em}
    \includegraphics[width=\linewidth]{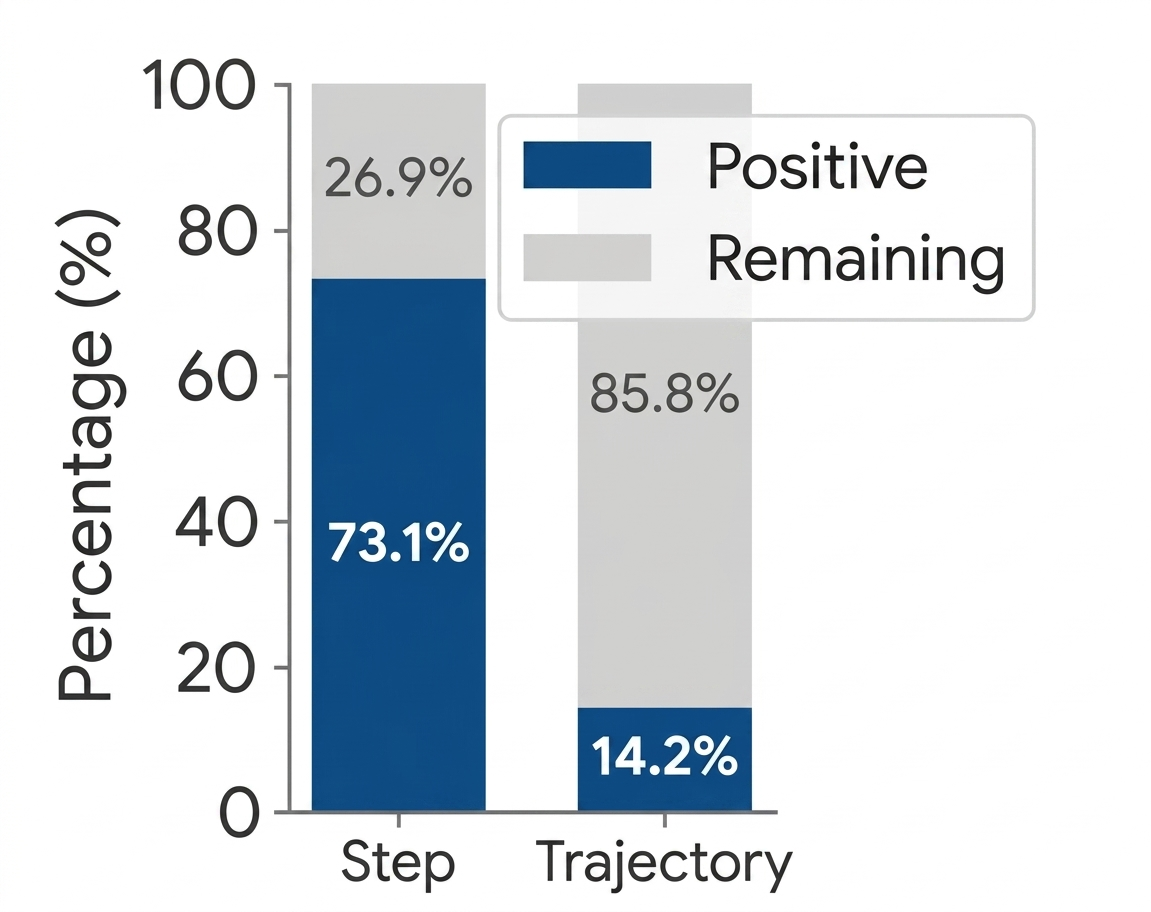}
    \caption{Step-level labels are substantially more skewed toward positives than trajectory-level labels in PRM800K.}
    \label{fig:data_imbalance}
    \vspace{-1.5em}
\end{wrapfigure}

\textbf{Data-Imbalance Problem.} We observed a key label imbalance in commonly available open-source PRM training datasets such as PRM800K \citep{lightman2023let}. PRM800K contains a much higher proportion of correct steps ($73.1\%$) than incorrect steps, even though only a small fraction of full trajectories are correct ($14.2\%$) (Figure~\ref{fig:data_imbalance}). There are two intuitive reasons behind this mismatch: (1) correct question--response pairs are easier to collect from open-source data, and (2) a trajectory is labeled incorrect as soon as even one step is wrong and a trajectory contains many steps. Thus, it can contain many locally correct steps while still being globally incorrect, making step-level labels far more skewed toward positives than trajectory-level labels.

\begin{table}[t]
    \centering
    \vspace{-0.6em}
    \resizebox{0.5\columnwidth}{!}{
    \begin{tabular}{lcccc}
    \toprule
    \textbf{Model} & \textbf{Pos-Acc/TPR} & \textbf{Neg-Acc/TNR} & \textbf{PRMScore} \\
    \midrule
    Qwen-PRM-7B & 95.36 & 30.66 & 65.5 \\
    ReasonEval-7B* & 95.5 & 21.2 & 60.0 \\
    ThinkPRM-7B & 83.29 & 50.89 & 64.3 \\
    GenPRM-7B & 52.25 & 73.92 & 50.5 \\
    GPT-4o* & 82.9 & 58.2 & 66.8 \\
    \bottomrule
    \end{tabular}
    }
    \caption{Positive and negative step accuracy for SOTA PRMs, along with PRMScores from \cite{song2025prmbench}.}
    \label{tab:intro_SOTA_PRMs}
    \vspace{-2.0em}
\end{table}

\textbf{Cross-Entropy Loss Exacerbates Bias.} In practice, PRMs are typically trained with pointwise, step-level cross-entropy loss \cite{zhang2025lessons, wang2023math}. Under label imbalance, such pointwise objectives are especially prone to bias. As Table~\ref{tab:intro_SOTA_PRMs} shows, existing models consistently over-predict correct steps. This stark false positive (FP) bias assigns spuriously high rewards to flawed reasoning, directly fueling downstream reward hacking.


PRM bias hurts downstream performance in two ways. In BoN and policy optimization, one overcredited step can raise the score of the entire response. In guided beam search, the effect is even stronger: an early false positive can push decoding onto a flawed path. In both cases, errors can accumulate over long chains, and the misalignment can grow with the number of steps, $T$.
\vspace{-2mm}
\subsection{How PRM Bias Distorts Downstream Policy Learning}\label{problem}
\vspace{-2mm}
PRMs are ultimately used to guide downstream policy optimization and search, so their quality should be evaluated through their downstream effect. Existing works focuse on standalone PRM accuracy, while overlooking its impact on policy learning and inference-time search. We therefore adopt a policy-aware view of PRM evaluation.

We illustrate how PRM bias affects downstream optimization through a standard KL-regularized RL objective. Without loss of generality, the same dependence also governs other downstream policy optimization (GRPO, PPO) and search tasks (BoN). The policy $\pi$ is optimized via:
\begin{equation}
\max_{\pi} \mathbb{E}_{y_t \sim \pi(\cdot|x, y_{<t})} [r_{\theta}(x, y_{<t}, y_t)] - \beta \mathbb{D}_{\text{KL}}(\pi(\cdot|x, y_{<t}) \parallel \pi_{\text{ref}}(\cdot|x, y_{<t}))
\label{eq:rlhf_objective}
\end{equation}
The optimal policy $\pi^*$ has the well-known closed-form solution:
\begin{equation}
\pi^*(y_t|x, y_{<t}) = \frac{\pi_{\text{ref}}(y_t|x, y_{<t}) \exp(r_{\theta}(x, y_{<t}, y_t)/\beta)}{Z(x, y_{<t})}
\label{eq:closed_form}
\end{equation}
where $Z(x, y_{<t})$ is the partition function. Equation~\ref{eq:closed_form} highlights a critical vulnerability: the exponentiated reward $\exp(r_{\theta}/\beta)$ makes the policy highly sensitive to overestimation. In particular, if a flawed step receives a spuriously high reward $r_{\theta}$ (a false positive, or FP), its probability under $\pi^*$ can become disproportionately large, making the optimizer learn incorrect information. Thus, reducing FPs is important to train a correct policy.

\subsubsection{Analyzing Asymmetric effect of PRM's False Positive Rate on Policy Search}\label{sec:theory} 
In this section, we study how PRM's false positives affect downstream policy search. In particular, we analyze the asymmetric effects of false positives and false negatives on Best-of-$N$ performance.




\begin{theorem}[Asymmetric Effect of False Positives (FPs) and False Negatives (FNs)]
\label{thm:fp_fn_asymmetry}
Consider Best-of-$N$ (BoN) selection using an imperfect binary reward model $\hat{T}(y) \in \{ 0,1 \}$ to select from $N$ i.i.d.\ samples $y_i \sim \pi(\cdot\mid x)$. Let $T(y)\in\{0,1\}$ be the true label, with base rate $p=\mathbb{P}(T=1)$. Let $\alpha=\mathbb{P}(\hat{T}=1\mid T=0)$ and $\beta=\mathbb{P}(\hat{T}=0\mid T=1)$ denote the FPR and FNR, respectively. Let $P^{(N)}$ be the probability BoN selects a response where $T=1$. Assuming $N$ is sufficiently large and the marginal predicted-positive rate is non-vanishing, then:

Asymmetry I: False positives induce a hard ceiling. If $\alpha>0$ and $\beta=0$, then:
\begin{equation}
\lim_{N\to\infty} P^{(N)} \;=\; \frac{p}{p+(1-p)\alpha} \;<\; 1
\label{eq:fp_ceiling}
\end{equation}
This ceiling is strictly decreasing in $\alpha$.

Asymmetry II: False negatives only slow convergence. If $\alpha=0$ and $\beta<1$, then:
\begin{equation}
\lim_{N\to\infty} P^{(N)} \;=\; 1
\label{eq:fn_convergence}
\end{equation}
In this case, FN reduce the rate at which $P^{(N)}$ approaches $1$, but do not induce asymptotic bias.

\end{theorem}

\begin{wrapfigure}{r}{0.38\textwidth}
    \centering
    \vspace{-0.5em} 
    \includegraphics[width=\linewidth, trim={0 0 0 0}, clip]{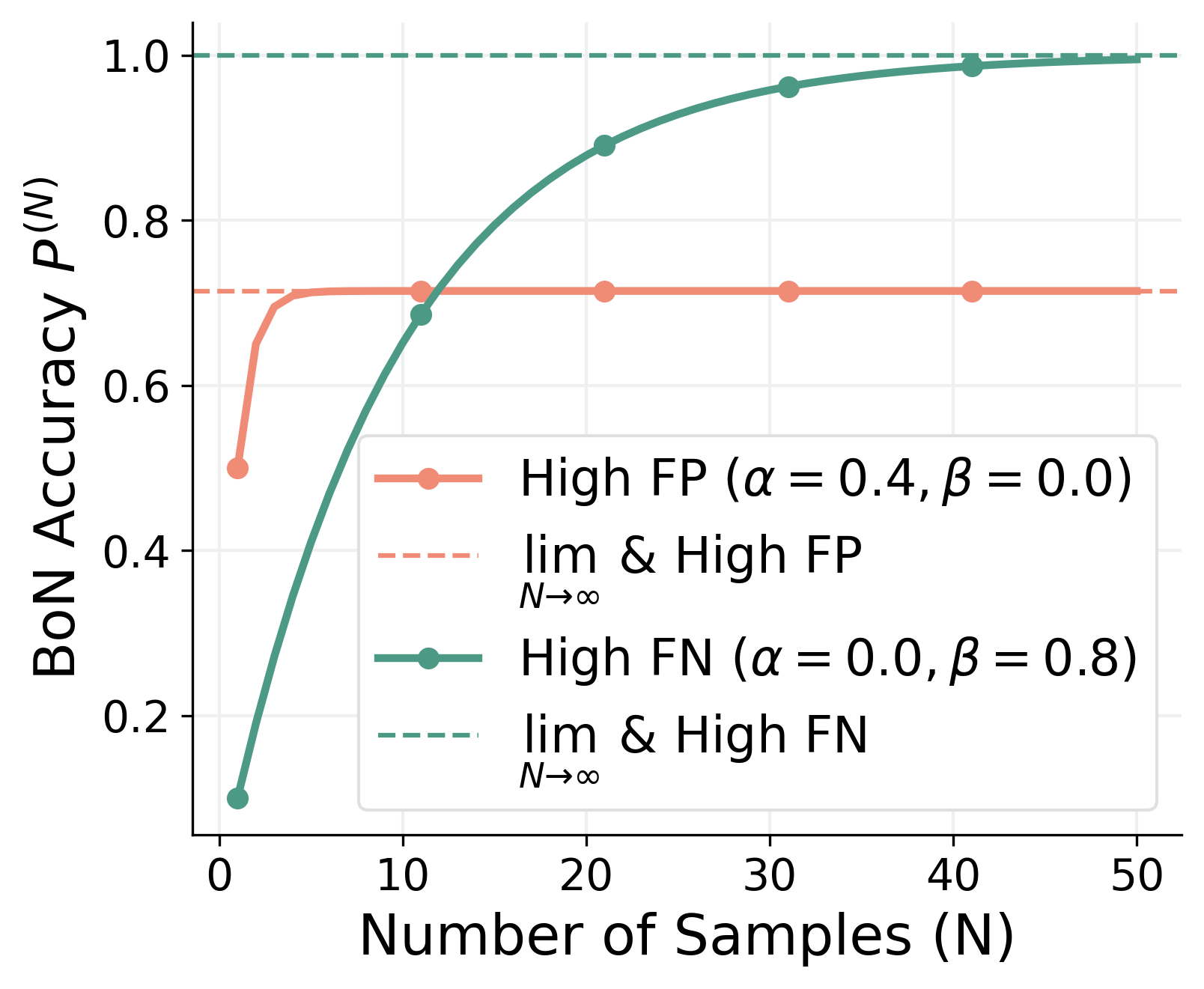}
    \caption{Effect of FP vs.\ FN on Bo$N$ Accuracy. FP induces a strict performance ceiling, while FN only slows the rate of convergence.}
    \label{fig:FP_FN}
    \vspace{-3.2em} 
\end{wrapfigure}



Thus, FN merely delays convergence, whereas FP fundamentally caps final performance under BoN section. Therefore, objective is to prioritize minimizing $\alpha$ (reducing FPs) to lift the ceiling on alignment. Detailed proof is provided in appendix section \ref{app:fp_fn_asymmetry}.

\vspace{-2mm}
\paragraph{Key Insight: Reducing False Positives Improves Policy Learning.} For PRMs, small overcredits can compound across long reasoning chains for guided decoding. In BoN and policy optimization, a single overcredited step can inflate the score of an incorrect response. As a result, a high FPR can misalign the policy and cause reward hacking. Objectives and data curation should therefore (i) rebalance step label distributions, and (ii) emphasize hard negatives, all aimed at minimizing $\alpha$ to lift the ceiling on alignment.

\vspace{-2mm}
\section{Policy Aware PRM Training to Mitigate PRM Bias}
\vspace{-2mm} 


There have been relatively few works on training and improving PRMs. However, these methods do not explicitly focus on downstream policy learning, which is the main setting in which PRMs are ultimately used. They also rely on mean-square-error loss \citep{chen2024alphamath} or pointwise losses \citep{wang2023math, luo2024improve, zhang2024rest} trained on human annotations \citep{lightman2023let} or MCTS-generated targets \citep{zhang2024rest, chen2024alphamath}. As a result, the precision of step-level annotations becomes critically important, yet remains difficult to obtain. In this work, we argue that PRMs should be evaluated through their effect on downstream policy learning. We therefore propose PRISM (Precision Ranking for Improved Step Modeling), a policy-aware and label-efficient framework for PRM training that combines a Step-Contrastive (SC) loss, hard-negative augmentation using a temporal lookahead strategy that requires no new human labels, and a difficulty-aware curriculum.



We first motivate our policy-aware approach to mitigating PRM bias, then introduce the loss, data augmentation strategy, and curriculum design. Note that, the goal of this work is not to design a perfect reward model but to highlight, and mitigate, systematic PRM failure modes that undermine alignment.


\vspace{-2mm}
\subsection{Step-Contrastive Loss}
\vspace{-2mm}
Here we motivate why we chose our proposed objective to train the PRM model followed by the exact details of our loss function. 
As discussed in section \ref{problem}, FPs can drive the policy toward incorrect reasoning. However, the goal of policy learning is simply to ensure that $\pi^*$ assigns higher probability to a correct step $y_t^{\text{pos}}$ than to an incorrect step $y_t^{\text{neg}}$. From Equation~\ref{eq:closed_form}, this relative preference can be written as:
\begin{equation}
\frac{\pi^*(y_t^{\text{pos}}|x, y_{<t})}{\pi^*(y_t^{\text{neg}}|x, y_{<t})} = \frac{\pi_{\text{ref}}(y_t^{\text{pos}}|x, y_{<t})}{\pi_{\text{ref}}(y_t^{\text{neg}}|x, y_{<t})} \exp \left( \frac{r_{\theta}(x, y_{<t}, y_t^{\text{pos}}) - r_{\theta}(x, y_{<t}, y_t^{\text{neg}})}{\beta} \right)
\label{eq:policy_ratio}
\end{equation}

Therefore, as long as $r_{\theta}(x, y_{<t}, y_t^{\text{pos}}) > r_{\theta}(x, y_{<t}, y_t^{\text{neg}})$, the exponentiated reward gap pushes the policy toward the correct step. The model does not need to separate the two steps by an arbitrarily large absolute amount; it only needs to rank them correctly. Standard binary cross-entropy (BCE), however, trains each step independently and does not optimize this relative comparison. Under label imbalance, it also becomes harder to keep hard negative steps sufficiently low-scoring. This motivates directly optimizing the margin which can be written as 
\[
m_\theta(x, y_{<t}) = r_\theta(x, y_{<t}, y_t^{\text{pos}}) - r_\theta(x, y_{<t}, y_t^{\text{neg}}).
\]
Thus, improving downstream policy learning reduces to increasing $m_\theta$. This directly motivates our step-contrastive loss which optimizes this margin. We introduce it below, 

\textbf{How we get step-scores.} Conventional PRM output reward, $r_{\theta}(\cdot)\in\mathbb{R}^{2\times 1}$, corresponding to scores for positive and negative labels. We use reward score with respect to positive class and call it $r_{\theta}(.)$.

\textbf{Step-Contrastive (SC) loss.}
We optimize a preference based objective, inspired from pairwise preference loss \cite{ouyang2022training}, that pushes the positive above the negative:
\vspace{-2mm}
\begin{equation}
\label{eq:oc_prm_loss}
\mathcal{L}_{\text{SC}}(\theta) = - \mathbb{E}_{(x,\, y_t^{\text{pos}},\, y_t^{\text{neg}})\in \mathcal{D}}
\Big[
\log\!\big(
\sigma\!\big(
r_\theta(x, y_{<t}, y_t^{\text{pos}}) - r_\theta(x, y_{<t}, y_t^{\text{neg}})
\big)
\big)
\Big],
\end{equation}
where $t \leq T$ can be any step uptil total steps $T$. Here $r_\theta(x, y_{<t}, y_t)$ is the PRM output for step $y_t$ given problem $x$ and prefix $y_{<t}$. Note, each pair $(y_t^{\text{pos}}, y_t^{\text{neg}})$ share the same prefix $(x, y_{<t})$.

Thus writing $\mathcal{L}_{\text{SC}}(\theta) = -\log \sigma(m_\theta)$ and taking its gradient we have,
\[
\triangledown_{\theta} \mathcal{L}_{\text{SC}}(\theta) = -\sigma(-m_\theta)\, \triangledown_{\theta} m_\theta.
\]
Since $\sigma(-m_\theta) > 0$, minimizing $\mathcal{L}_{\text{SC}}$ pushes $\theta$ in the direction that increases $m_\theta$. In simple terms, the objective directly encourages the reward of a correct step to exceed that of an incorrect step.

\textbf{What this buys us.}
Because $(x, y_{<t})$ is shared, the model learns a \emph{relative} preference within the same context. The objective places the largest penalty when an incorrect step outranks, or nearly ties, a correct step—directly reducing overcredit on negatives. It is architecture agnostic and uses only existing step labels.


\paragraph{Why Step-Contrastive Loss Is Better Aligned Than Step-BCE For Policy Learning} 
\vspace{-1mm}

\begin{wrapfigure}{r}{0.3\linewidth}
    \centering
    \vspace{-1.2em}
    \includegraphics[width=\linewidth]{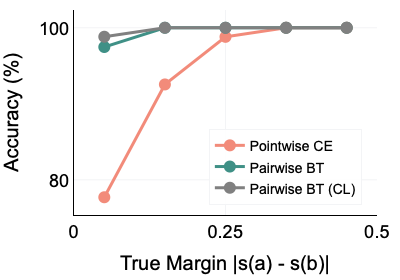}
    \caption{Pairwise classification accuracy of Pointwise versus Pairwise models.}
    \label{fig:confusion_region}
    \vspace{1.2em}
\end{wrapfigure}

To show this, we conducted simulation by training two identical network architectures using CE loss optimized to predict absolute binary labels, and a BT loss optimized directly on pairwise preferences on the same matched dataset $\mathcal{D}=\{(x, y_{<t}, y_t^{\text{pos}}, y_t^{\text{neg}})\}$. We show pairwise classification accuracy for different absolute margin $|s(a) - s(b)| \in [0, 0.5]$. As shown in Figure \ref{fig:confusion_region}, the BT model maintains significantly higher accuracy as the margin approaches zero. This is because BCE optimizes the two steps independently rather than directly maximizing their contextual margin. In contrast, pairwise objective learns a relative preference between two steps, which is better aligned with downstream task. For more details check Appendix \ref{appx:explain_pair_point}.


Training with the step-contrastive loss requires matched positive--negative pairs with hard negatives that need no new human labels. We discuss this in section \ref{sec:aug}. We also need a curriculum strategy for learning a stronger margin as discussed in section \ref{sec:curr}.  
\vspace{-1mm}

\subsection{Hard-Negative Augmentation Using Temporal Lookahead}\label{sec:aug}
\vspace{-2mm}

To train with the step-contrastive loss, we need matched positive--negative step pairs whose negatives are genuinely hard: plausible continuations that can confuse the reward model while remaining invalid for the current prefix. Obtaining reliable annotations at this granularity is difficult, so we first use PRM800K's positive, negative, and neutral labels to form explicit positive--negative pairs from annotated intermediate steps, yielding about 26k training pairs. We then apply temporal lookahead to create additional hard negatives.

Simpler augmentation alternatives do not satisfy these hard-negative constraints as cleanly. Random negatives from other problems are often too easy because they are contextually unrelated; synthetic perturbations can introduce superficial artifacts that the model may exploit; and model-generated adversarial negatives require extra decoding, filtering, or human validation.

Our temporal lookahead augmentation strategy instead treats a future step from the same trajectory as a negative candidate for the current prefix. This teaches the model to judge whether a step is correct \emph{at the current point} in the reasoning chain, not merely whether it is valid in isolation. The resulting pairs are challenging: the future step is usually relevant to the problem and may even be correct on its own, but it is still an invalid next step because it appears too early. For example, in a quadratic-equation solution, writing the final root before the intermediate simplifications may be mathematically valid, but it is not a valid continuation of the current partial solution. After augmentation, we obtain roughly 220k training pairs without any additional human labeling.
\vspace{-1mm}
\subsection{Margin Difficulty-Aware Curriculum Design}\label{sec:curr}

\vspace{-1mm}
Learning a strong margin from hard positive--negative pairs with a contrastive loss is often noisy for the model. This is because pairwise losses are known to struggle on overly difficult comparisons \citep{gao2025principled, wu2024towards}. 

A natural strategy is therefore to gradually improve the margin by progressively introducing harder positive--negative pairs during training. We design a novel curriculum learning \citep{inproceedings} paradigm in which the margin itself determines the difficulty level. Specifically, we define the continuous difficulty score as the difference between the reward assigned to the positive and negative step. Let us call this score:
$$D_{\text{Hard}} = r_\theta(x, y_{<t}, y_t^{pos}) - r_\theta(x, y_{<t}, y_t^{neg})$$
where $0 \leq r_\theta(x, y_{<t}, y_t) \leq 1$. Thus, $D_{\text{Hard}} \in [-1, 1]$. We retain pairs with $D_{\text{Hard}} \geq 0$ and discard the rest, since negative values typically correspond to comparisons that are too difficult for the model to learn from reliably. The retained pairs are then divided into four bins; the number of samples in each bin is reported in Appendix~\ref{appx:CL_samplenum}. We also compare training with and without curriculum learning on the same set of training examples in Tables~\ref{tab:PRMbench_results}, and provide ablations over different curriculum bins to show robustness to this hyperparameter. 

Since existing PRMs, both discriminative and generative, exhibit high FPR and are trained on imbalanced class-label data, our proposed loss is broadly applicable to both.

%% file: sections/4.Results_Experiments.tex
\vspace{-2mm}
\section{Experiments and Results}
\vspace{-1mm}

\paragraph{Models and Baselines.} The SOTA PRM model Qwen2.5-Math-PRM-7B \citep{zhang2025lessons} serves as the primary baseline for all experiments. To test generalizability, ReasonEval-7B PRM \citep{xia2025evaluating} is also included in select experiments. Models trained with the proposed recipe are denoted as \textbf{PRISM} hereafter. Because Qwen-PRM is a discriminative model, comparisons additionally include ThinkPRM \citep{khalifa2025process}, a SOTA generative PRM. A standard baseline trained on the same dataset with the default pointwise loss is included as well.

\textbf{PRM Evaluation Benchmarks.} Step-level reasoning quality is evaluated using \textbf{PRMBench} \citep{song2025prmbench} and \textbf{ProcessBench} \citep{zheng2025processbench}. PRMBench assesses the correctness, faithfulness, and robustness of intermediate chain-of-thought steps, providing detailed step-level accuracies and confusion matrices.

\begin{table}[t]
\centering
\resizebox{\textwidth}{!}{%
\begin{tabular}{l ccccccc}
\toprule
\textbf{Method} & \textbf{FPR} $\downarrow$ & \textbf{FNR} $\downarrow$ & \textbf{Precision} $\uparrow$ & \textbf{TPR} $\uparrow$ & \textbf{TNR} $\uparrow$ & \textbf{PRMScore} $\uparrow$ & \textbf{FPR Reduction} $\uparrow$ \\
\midrule
Baseline (Qwen-PRM-7B) & 69.34 & 4.27 & 89.40 & 95.36 & 30.66 & 65.50 & 0.00\% \\
\midrule
\quad + Pairwise Loss (12k data) & 61.13 & 7.10 & 88.96 & 92.41 & 38.86 & 67.00 & 11.84\% \\
\quad + Pairwise Loss (12k data) + CL & 60.58 & 7.20 & 90.4 & 92.79 & 39.42 & 67.30 & 12.63\% \\
\midrule
\quad + Pointwise Loss + Data Aug. (132k) & 67.00 & 6.00 & 89.60 & 94.00 & 33.50 & 65.60 & 3.37\% \\
\textbf{PRISM} & \textbf{47.13} & \textbf{12.69} & \textbf{91.93} & \textbf{87.30} & \textbf{52.86} & \textbf{68.00} & \textbf{32.22\%} \\ 
\bottomrule
\end{tabular}%
}
\vspace{2mm}
\caption{Reported metrics include the False Positive Rate (FPR), False Negative Rate (FNR), Precision, TPR, TNR, and overall PRMScore. Because global accuracy is distorted by heavy class imbalance, these localized metrics better reflect model alignment. \textbf{PRISM, trained via pairwise loss, data augmentation, and curriculum learning,} significantly reduces FPR while improving overall PRMScore. Note: A 1--2\% increase in PRMScore translates to substantial downstream capability gains \citep{song2025prmbench}.}
\vspace{-4mm}
\label{tab:pos_neg_accuracy}
\end{table}

\begin{figure}[hbt!]
    \centering
    \begin{subfigure}[b]{0.5\textwidth}
        \centering
        \includegraphics[width=\textwidth, trim={0 0 0 0}, clip]{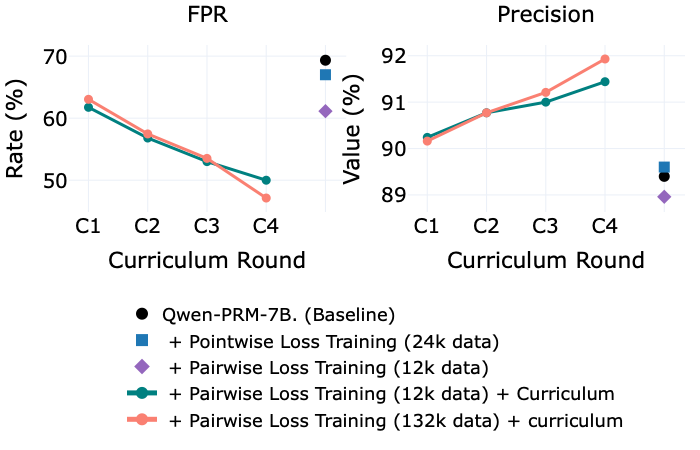}
        \caption{FPR and precision across curriculum rounds.}
        \label{fig:CL_rounds_a}
    \end{subfigure}
    \hfill 
    \begin{subfigure}[b]{0.49\textwidth}
        \centering
        \includegraphics[width=\textwidth, trim={0 0 0 0}, clip]{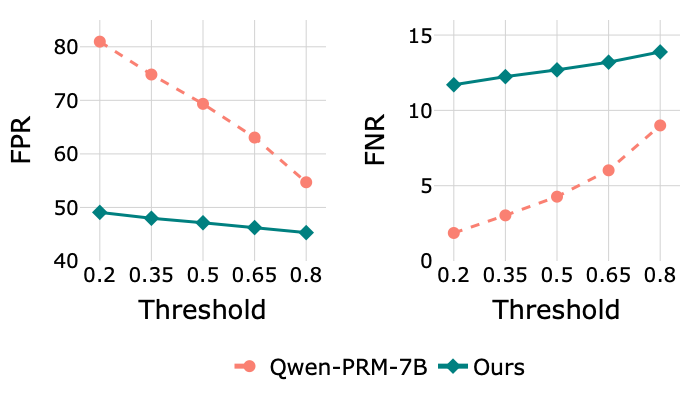}
        \caption{FPR--FNR trade-off across thresholds.}
        \label{fig:thresholding}
    \end{subfigure}
    \caption{Curriculum and threshold analyses for PRISM. \textbf{Left:} Later curriculum rounds reduce false positives while maintaining strong precision. \textbf{Right:} Across classification thresholds, PRISM achieves a more favorable FPR--FNR trade-off than the baseline, indicating greater robustness to threshold selection.}
    \vspace{-4mm}
    \label{fig:CL_rounds_combined}
\end{figure}

\textbf{ProcessBench} targets complex mathematical reasoning, evaluating a model's ability to pinpoint the first erroneous step in Olympiad-level solutions.

\textbf{Policy Evaluation.} Alignment is evaluated at both inference and training time. For \textbf{inference-time evaluation}, \textbf{best-of-$N$} and \textbf{guided beam-search} experiments are conducted using both in-distribution (ID) and out-of-distribution (OOD) policies. Evaluation covers MATH-500 \citep{hendrycks2021measuring}, AIME 2024 problems, and LiveCodeBench \citep{jain2024livecodebench}. Following \cite{khalifa2025process}, 100 MATH-500 problems spanning all difficulty levels are sub-sampled. For ID policies, the Qwen family is used: Qwen2.5-Math-1.5B-Instruct for MATH-500, Qwen2.5-Math-7B-Instruct for AIME 2024, and Qwen2.5-Coder-7B-Instruct \citep{hui2024qwen2} for coding tasks. For OOD evaluation on MATH-500, Llama-3.2-3B-Instruct is used. Finally, for \textbf{training-time evaluation}, policies are optimized using the GRPO algorithm on the Math-Verify dataset.

\subsection{PRM Evaluation}

\paragraph{PRMBench: Lower FPR and Higher Precision with PRISM.} Table~\ref{tab:PRMbench_results} presents detailed PRMBench results for PRISM. Additionally, Table~\ref{tab:pos_neg_accuracy} reports step-level metrics---including positive and negative accuracy, false-positive rate (FPR), false-negative rate (FNR), and precision---to evaluate the reduction in false positives.

A significant reduction in FPR is observed alongside a slight increase in precision, indicating that the model is more conservative in its positive predictions. While this is accompanied by a marginal rise in FNR, this degradation is substantially smaller than the FPR improvement. Moreover, unlike FPR, FNR does not directly harm policy alignment. Overall, the model demonstrates consistent performance gains. Although the numerical increase in PRMScore appears modest, such improvements translate to substantial gains in downstream model capability~\citep{song2025prmbench}.

\textbf{FPR-FNR Trade-off Across Classifier Thresholds:} Figure~\ref{fig:thresholding} illustrates the FPR-FNR curves for various classifier thresholds. PRISM demonstrates significantly greater robustness to threshold variations compared to the baseline.\\
\textbf{Sensitivity to Curriculum Learning Bins:} Ablation results across different curriculum learning bins (Appendix~\ref{appx:cl_bins_ablation}) consistently reflect this same performance trend as discussed before.\\
\textbf{Generalization to ReasonEval-7B PRM:} Results in Appendix~\ref{appx:reasoneval_results} demonstrate that PRISM generalizes effectively beyond Qwen-PRM.\\
\textbf{ProcessBench Evaluation:} Evaluations on ProcessBench~\citep{zheng2025processbench} (Appendix~\ref{appx:processbench}) reveal trends consistent with the PRMBench findings.

\vspace{-2mm}
\subsection{Policy Evaluation: Downstream Alignment}
\vspace{-1mm}
\subsubsection{Inference-Time Search Evaluation}
\vspace{-1mm}

To evaluate whether PRM improvements translate into better policy alignment, both in-distribution (ID) and out-of-distribution (OOD) policies (generators) are tested using two alignment algorithms: \textbf{guided beam-search} and \textbf{best-of-N}. For both methods, a verifier-weighted majority rule is used to select the final answer, where answers are scored based on the sum of verifier scores across their supporting solutions \citep{wu2024empirical, uesato2022solving}. A detailed description of this selection strategy is provided in Appendix~\ref{appx:answer_selection}.
Results are reported using two generator model families: Qwen (ID) and LLaMA (OOD). Using multiple policy models is important because PRM bias is especially likely to surface when either the prompt or the response distribution is OOD relative to the PRM. OOD policies are especially likely to produce OOD responses, making them a strong testbed for alignment performance. Evaluating across different generators also ensures that the findings are not tied to a particular model family or size. \\
\textbf{Guided Beam-Search.} This is an extension of standard beam search. It incorporates verifier (PRM) scores when ranking partial reasoning chains. Instead of relying solely on model likelihoods, the search is guided toward trajectories that both the model and the verifier find promising, improving alignment with correct reasoning. \\
\textbf{Best-of-N.}  samples $N$ candidate solutions from the policy and then uses the PRM (or verifier) to score each solution. \\

\begin{figure}[!ht]
    \centering
    \includegraphics[width=\linewidth, trim={0 40 0 40}, clip]{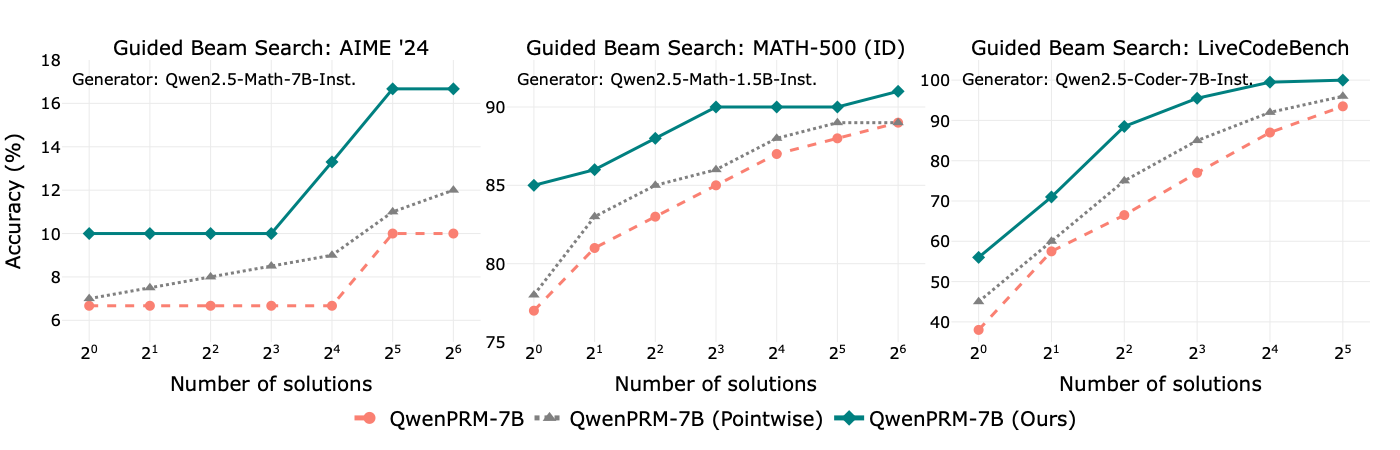}
    \vspace{1em}
    \includegraphics[width=\linewidth, trim={0 0 0 20}, clip]{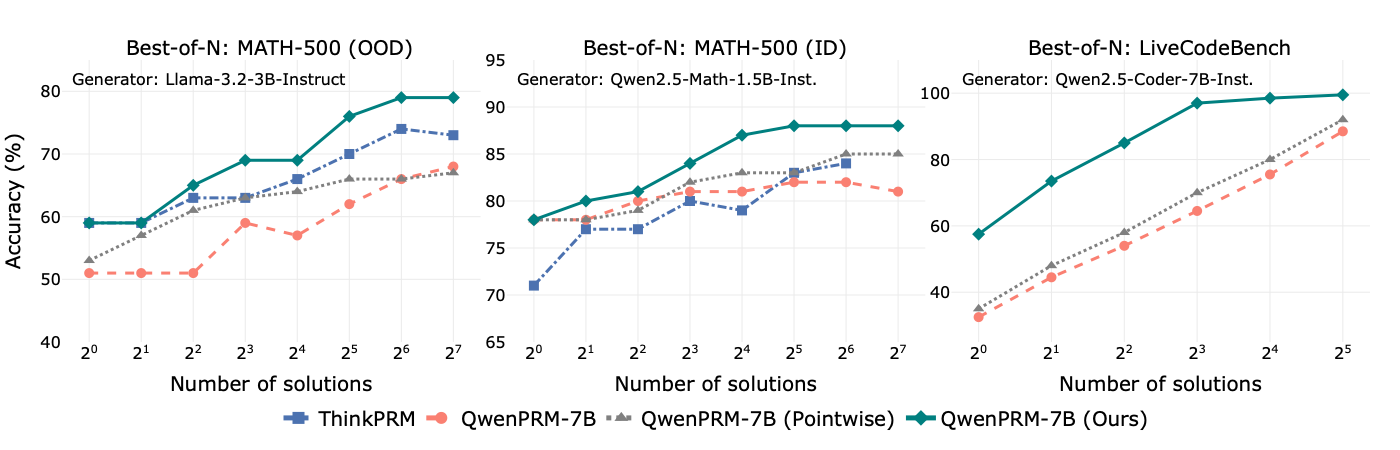}
    \vspace{-2em}
    \caption{Comparison of Best-of-$N$ and guided beam-search alignment on MATH-500, AIME’24, and LiveCodeBench. \textbf{Top:} Guided beam-search results on AIME’24 (left), MATH-500 (middle), and LiveCodeBench (right). Across all settings, PRISM (blue) consistently outperforms the baseline (orange), highlighting its effectiveness in guided beam-search decoding, where the PRM plays a critical role.
    \textbf{Bottom:} Best-of-$N$ results on MATH-500 using two generator policies, LLaMA for OOD (left) and Qwen for ID setting (middle), and on LiveCodeBench using the Qwen2.5-Coder generator (right). ThinkPRM is shown in gray. PRISM achieves clear performance gains across all settings, with especially large improvements in the OOD regime.
    }
    \label{fig:guided_beam_search}
    \vspace{-3mm}
\end{figure}

\textbf{Improvement in Alignment Performance.} 
Figures~\ref{fig:guided_beam_search} and 
\ref{fig:guided_OOD} show alignment performance for both guided beam search and best-of-N strategies on MATH-500, AIME and LiveCodeBench tasks. Consistent and significant improvements appear across all settings. Notably, guided beam search exhibits substantially larger gains with PRISM than with the baselines, indicating a much stronger alignment signal and suggesting that PRISM is more effective. BoN relies on random sampling from the policy, after which the PRM selects the best solution from a fixed set. In contrast, guided beam search actively shapes generation by incorporating PRM feedback at each decoding step. This forces the PRM to reliably distinguish promising partial trajectories, providing a stronger and more realistic evaluation of whether the PRM captures true step-level reasoning quality rather than simply identifying winners post hoc. \\
\textbf{Improvement in Alignment for both ID and OOD policy.} Evaluating OOD policies is particularly important for studying bias-induced policy misalignment, since reward models are most vulnerable when they encounter data outside their training distribution. In such cases, the model may assign high rewards to spurious or incorrect reasoning. OOD policies naturally produce more OOD responses, making them an effective stress test for alignment robustness. Results in Figures~\ref{fig:guided_beam_search} and \ref{fig:guided_OOD} show that PRISM reduces bias-induced policy misalignment for both ID and OOD policies, confirming stronger generalization. \\
\textbf{Improvement in Alignment across diverse datasets and tasks:} Both mathematics and coding tasks are included. For mathematics, evaluation covers MATH-500 and AIME’24, where MATH-500 is closer in difficulty to the paired training data while AIME’24 is more challenging. Improvements on MATH-500 confirm that PRISM strengthens alignment within distribution, while gains on AIME’24 demonstrate generalization to more difficult, out-of-distribution problems. For coding, evaluation uses LiveCodeBench \citep{jain2024livecodebench}. These results, shown in Figure~\ref{fig:guided_beam_search}, indicate that the method improves the model’s underlying reasoning ability rather than simply overfitting to training-like distributions.

\vspace{-2mm}
\subsubsection{Training-Time Policy Optimization via GRPO}
\vspace{-2mm}

To assess practical impact, a Qwen2.5-1.5B policy is optimized using GRPO. As shown in Table~\ref{tab:grpo_results}, substituting the Qwen-PRM-7B baseline with PRISM yields significant downstream gains. Most notably, GSM8K improves by an absolute $+11.37\%$, alongside strong gains on AMC and MATH-500. These results demonstrate that optimizing the FPR-FNR trade-off directly translates to a more capable generative policy.

\begin{table}[hbt!]
\centering
\resizebox{0.8\textwidth}{!}{%
\begin{tabular}{l cccccc}
\toprule
\textbf{PRM Model} & \textbf{AIME} & \textbf{AMC} & \textbf{GSM8K} & \textbf{MATH-500} & \textbf{Minerva} & \textbf{Olympiad} \\
\midrule
Baseline (Qwen-PRM-7B) & 0.00 & 3.61 & 31.16 & 14.20 & \textbf{6.25} & \textbf{5.04} \\
\textbf{PRISM (Ours)} & 0.00 & \textbf{9.64} & \textbf{42.53} & \textbf{15.80} & 5.88 & 4.00 \\
\bottomrule
\end{tabular}%
}
\vspace{2mm}
\caption{Downstream GRPO performance on a Qwen2.5-1.5B policy. Using PRISM as the reward signal drives substantial gains on core mathematical reasoning benchmarks compared to the baseline.}
\label{tab:grpo_results}
\end{table}

%% file: sections/5.RelatedWorks.tex
\vspace{-6mm}
\section{Related Work}
\vspace{-2mm}

\textbf{Process Reward Models.} There are two kinds of PRM: Discriminative PRMs and Generative PRMs. Discriminative PRMs are typically framed as classification tasks, where the model assigns a correctness score to each reasoning step. These models require step-level supervision \citep{uesato2022solving, lightman2023let, zhang2025lessons}. For a given solution prefix, the text is encoded and passed through a classification head that outputs step-level correctness probabilities, commonly trained using binary cross-entropy loss. To evaluate a full solution, the step-level scores are aggregated into an overall correctness measure \citep{beeching2024scalingtesttimecompute, snell2024scaling, wu2024empirical}. Generative process reward models (PRMs) \citep{khalifa2025process, zhao2025genprm, zheng2023judging, zhu2023judgelm} treat verification as a sequence generation problem, where the model outputs natural language tokens such as “correct” or “incorrect” at each reasoning step. Instead of relying solely on binary labels, they are trained with the standard language modeling objective using explanatory rationales. 

\textbf{PRM Bias and Reward-Model Robustness}
Reward hacking is the broader term typically used in reinforcement learning \citep{skalse2022defining}, and closely related reward-model bias has also been explored extensively in the context of ORMs for LLMs. However, little to no work has addressed PRM bias directly. Prior works have used reward model uncertainty as a training signal to guide policy learning, which in turn generates OOD samples for improving the reward model \cite{bukharin2025adversarial}. Their approach, however, relies on the strong assumption that all OOD samples correspond to incorrect outputs. Other lines of work have similarly proposed reward ensembles and Bayesian methods to improve robustness and reduce vulnerability to reward-model bias and proxy misalignment \citep{yan2024reward, yang2024bayesian}. Beyond Bayesian approaches, \cite{liu2024rrm} employ counterfactual augmentations to create paired data, explicitly breaking label-specific artifacts that might otherwise mislead the reward model. Even though the broader reward-hacking literature has not yet focused much on PRMs, shortcut behaviors—closely related to PRM bias—are pervasive in reasoning LLMs \citep{baker2025monitoring, denison2024sycophancy}, which are often trained with process supervision. This highlights the importance of explicitly investigating and addressing PRM bias in process reward models.

%% file: sections/6.Conclusion_Limitation.tex
\vspace{-4mm}
\section{Conclusion and Limitations}
\vspace{-2mm}

In this work we first identify and then characterize a hidden bias in PRMs: false-positive overcrediting, where plausible but incorrect reasoning steps receive undeservedly high rewards. We show that this overcredit bias is not merely an offline evaluation error, but a policy-level failure mode. False positives have an asymmetric downstream effect: unlike false negatives, which mainly slow exploration, they can actively steer Best-of-$N$ selection, guided decoding, and policy optimization toward flawed reasoning. We further show that imbalanced step-level data and pointwise objectives such as BCE can aggravate this bias, motivating a policy-aware view of PRM evaluation and training. To mitigate this failure mode, we propose \ours, a label-efficient PRM training recipe that replaces pointwise label fitting with step-contrastive learning, constructs hard negatives via temporal lookahead without new human annotations, and stabilizes training with a difficulty-aware curriculum. Across PRM benchmarks and downstream alignment tasks, this precision-focused design reduces false positives and improves both verifier quality and reasoning performance.

A key limitation remains the trade-off between false positives and false negatives: reducing false positives can sometimes increase false negatives, and achieving both high precision and high recall remains an important direction for future work. Developing PRMs that maintain this balance across diverse tasks, policies, and out-of-distribution reasoning traces is crucial for building more reliable reasoning systems.



%% file: sections/Appendix.tex
\onecolumn


\section{Software and Hardware Details}
All experiments were run on four Nvidia RTX A6000 GPUs.

\section{Theoretical Results and Insights}

\subsection{False Positives vs. False Negatives in Policy Misalignment}
\label{app:fp_fn_asymmetry}

This section analyzes how reward model errors (false-positive and false-negative prediction) affect Best-of-$N$ (BoN) selection.

\subsubsection[Setup: Best-of-N Under an Imperfect Reward Model]{Setup: Best-of-$N$ Under an Imperfect Reward Model}

Let $y_1,\dots,y_N \sim \pi(\cdot\mid x)$ be i.i.d.\ samples.
Let $T(y)\in\{0,1\}$ be the true binary label and $\hat{T}(y)\in\{0,1\}$ the reward model prediction.
Best-of-$N$ (BoN) selects
\begin{align}
y^* = \arg\max_{i\in\{1,\dots,N\}} \hat{T}(y_i).
\end{align}

Define
\begin{align}
p &= \mathbb{P}(T=1), \\
\alpha &= \mathbb{P}(\hat{T}=1 \mid T=0) \quad \text{(false positive rate)}, \\
\beta &= \mathbb{P}(\hat{T}=0 \mid T=1) \quad \text{(false negative rate)}.
\end{align}

The marginal probability that a sample is predicted positive is
\begin{align}
q = (1-p)\alpha + p(1-\beta).
\end{align}

Hence, the probability that at least one of the $N$ samples is predicted positive is
\begin{align}
\hat{p}_N = 1 - (1-q)^N.
\end{align}

Conditioned on $\hat{T}=1$, the precision of the reward model is
\begin{align}
\mathbb{P}(T=1 \mid \hat{T}=1) = \frac{p(1-\beta)}{q}.
\end{align}

Therefore, the BoN accuracy is
\begin{align}
P^{(N)} &=
\underbrace{\Bigl[1-(1-q)^N\Bigr]\mathbb{P}(T=1 \mid \hat{T}=1)}_{\text{predicted-positive event}}
\;+\;
\underbrace{(1-q)^N \mathbb{P}(T=1 \mid \hat{T}=0)}_{\text{all-negative event}} \\
&=
\Bigl[1-(1-q)^N\Bigr]\frac{p(1-\beta)}{q}
\;+\;
(1-q)^N
\frac{p\beta}{(1-p)(1-\alpha)+p\beta}.
\label{eq:bon_full}
\end{align}
In the regime of interest where $N$ is moderate to large and the predicted-positive rate $q$ is not vanishing, the probability of the all-negative event, $(1-q)^N$, decays exponentially in $N$. Consequently, the second term in~\eqref{eq:bon_full} contributes only a lower-order correction to $P^{(N)}$. For clarity of exposition, and to isolate the dominant mechanism governing BoN behavior, the leading term is used to approximate
\begin{align}
P^{(N)} = \hat{p}_N \cdot \mathbb{P}(T=1 \mid \hat{T}=1)
= \Bigl[1-(1-q)^N\Bigr]\frac{p(1-\beta)}{q}.
\end{align}

\subsubsection{Asymmetry I: False Positives Impose a Precision Ceiling}

If $\beta \approx 0$, then $q = p + (1-p)\alpha$ and
\begin{align}
P^{(N)} = \Bigl[1-(1-q)^N\Bigr]\frac{p}{p+(1-p)\alpha}.
\end{align}

Taking the limit $N\to\infty$ yields,
\begin{align}
\lim_{N\to\infty} P^{(N)} = \frac{p}{p+(1-p)\alpha} < 1 \quad \quad (\alpha>0).
\end{align}

Thus, nonzero false positives impose a permanent precision ceiling. Moreover,

\begin{align}
\frac{\partial}{\partial \alpha}
\left(\frac{p}{p+(1-p)\alpha}\right)
= -\frac{p(1-p)}{(p+(1-p)\alpha)^2} < 0,
\end{align}
showing that the ceiling degrades monotonically with $\alpha$. This can also be visualized in Figure \ref{fig:FP_FN}.

\subsubsection{Asymmetry II: False Negatives Slow but Do Not Cap Alignment}

If $\alpha \approx 0$, then $q = p(1-\beta)$ and
\begin{align}
P^{(N)} = 1 - \bigl(1-p(1-\beta)\bigr)^N.
\end{align}

Taking the limit $N\to\infty$ yields,
\begin{align}
\lim_{N\to\infty} P^{(N)} = 1.
\end{align}

Thus, false negatives reduce the effective success probability per sample but do not induce asymptotic bias; increasing $N$ fully compensates for false negatives.

\subsection{Pairwise vs. Pointwise Loss}
\label{appx:proof_pairwisevspointwise}

\paragraph{Setup.}
Let $s_t^{+} = r_\theta(x, y_{<t}, y_t^{\text{pos}})$ and $s_t^{-} = r_\theta(x, y_{<t}, y_t^{\text{neg}})$. For one pair, pointwise BCE optimizes the two scores independently,
$$\mathcal{L}_{\text{BCE}}(s_t^{+}, s_t^{-}) = -\log s_t^{+} - \log(1 - s_t^{-}),$$
whereas pairwise contrastive training optimizes only their relative ordering, 
$$\mathcal{L}_{\text{SC}}(s_t^{+}, s_t^{-}) = -\log \sigma(s_t^{+} - s_t^{-}).$$

\subsection{Why Pairwise Training Outperforms Pointwise Training}\label{appx:explain_pair_point}

The difference is easiest to see through a simple example. Consider a hard negative in mathematical reasoning: a step that sets up most of the solution correctly but makes a subtle arithmetic error at the end. Such a step shares substantial \textit{valid} reasoning with the correct step. Suppose the PRM assigns scores $0.85$ to the correct step and $0.80$ to the hard negative. For downstream policy learning, this is already mostly a success because the ranking is correct ($0.85 > 0.80$), so the policy prefers the correct step through the exponentiated margin. BCE, however, still treats the $0.80$ negative score as a large error and pushes it aggressively toward $0$. Because the hard negative shares much of its structure with the positive step, this can create gradient conflict by suppressing useful reasoning patterns and may increase FNs. Step-Contrastive loss instead focuses only on the margin $m=0.05$, applying a targeted update that widens the gap while allowing both steps to retain relatively high absolute scores.

The difference is even clearer when the ranking is reversed, for example $s_t^{+}=0.40$ and $s_t^{-}=0.45$. BCE again spends effort pushing the scores toward $1$ and $0$, whereas Step-Contrastive training focuses directly on flipping the ordering. This margin-centric view also motivates curriculum learning: large positive margins correspond to easy pairs, small positive margins correspond to hard but learnable pairs, and negative margins are often the most unstable early in training. Training is therefore organized from easier to harder pairs based on the predicted margin $s_t^{+} - s_t^{-}$, which stabilizes optimization, avoids unstable early gradients, and sharpens the decision boundary.

Writing the scores in center--margin coordinates
\[
a_t = \tfrac{1}{2}(s_t^{+} + s_t^{-}), \qquad m_t = s_t^{+} - s_t^{-}.
\]
Then
\[
\mathcal{L}_{\text{BCE}}(a_t,m_t) = -\log\!\left(a_t+\tfrac{m_t}{2}\right) - \log\!\left(1-a_t+\tfrac{m_t}{2}\right),
\qquad
\mathcal{L}_{\text{SC}}(a_t,m_t) = -\log \sigma(m_t).
\]
Equation~\ref{eq:policy_ratio} depends only on $m_t$, not on the center $a_t$. Accordingly,
\[
\frac{\partial \mathcal{L}_{\text{SC}}}{\partial a_t}=0, \qquad
\frac{\partial \mathcal{L}_{\text{SC}}}{\partial m_t}=-\sigma(-m_t)<0,
\]
so SC is an exact surrogate for increasing the policy-relevant margin. By contrast, BCE also has
\[
\frac{\partial \mathcal{L}_{\text{BCE}}}{\partial a_t}
=
-\frac{1}{a_t+m_t/2}
+
\frac{1}{1-a_t+m_t/2},
\]
which is generally nonzero. Thus, even on the same balanced paired data, BCE expends optimization effort on the absolute score level $a_t$, a nuisance quantity that cancels out in Equation~\ref{eq:policy_ratio}.

Now let $\tau$ be a decision threshold. Since
\[
\mathrm{FNR}(\tau)=\mathbb{P}(s_t^{+}<\tau)=\mathbb{P}\!\left(m_t < 2(\tau-a_t)\right),
\qquad
\mathrm{FPR}(\tau)=\mathbb{P}(s_t^{-}\ge\tau)=\mathbb{P}\!\left(m_t \le 2(a_t-\tau)\right),
\]
a rightward shift of the margin distribution $m_t$ lowers both the probability that negatives cross the threshold and the probability that positives fall below it, provided the center $a_t$ is not simultaneously shifted upward. Hence a loss that directly enlarges $m_t$ creates a wider threshold region in which FPR decreases without increasing FNR. 

\subsubsection{Distribution Shift: Pairwise vs. Pointwise Loss}
\label{appx:dist_plot_pair_point}

\textbf{Data Imbalance in Pointwise vs.\ Pairwise Loss.}
Class rebalancing can also be applied to pointwise (cross-entropy) training. However, pointwise loss trains a binary classifier that predicts the absolute correctness of an individual step (``is this step correct or not?''), whereas pairwise loss trains a model to learn relative preference between two steps (``is step A better than step B?''). The latter directly optimizes ranking quality, which downstream alignment methods rely on, as reflected in the BON scores. The margin induced by the pairwise loss also leads to a clearer separation between positive and negative steps, which is crucial for reducing false positives.

To illustrate this difference, the same Qwen-PRM is fine-tuned with both pairwise and pointwise losses on identical positive and negative data. Table~\ref{tab:fpr_comparison} reports the resulting false-positive rate (FPR) comparison.

\begin{figure}
    \centering
    \includegraphics[width=0.4\linewidth]{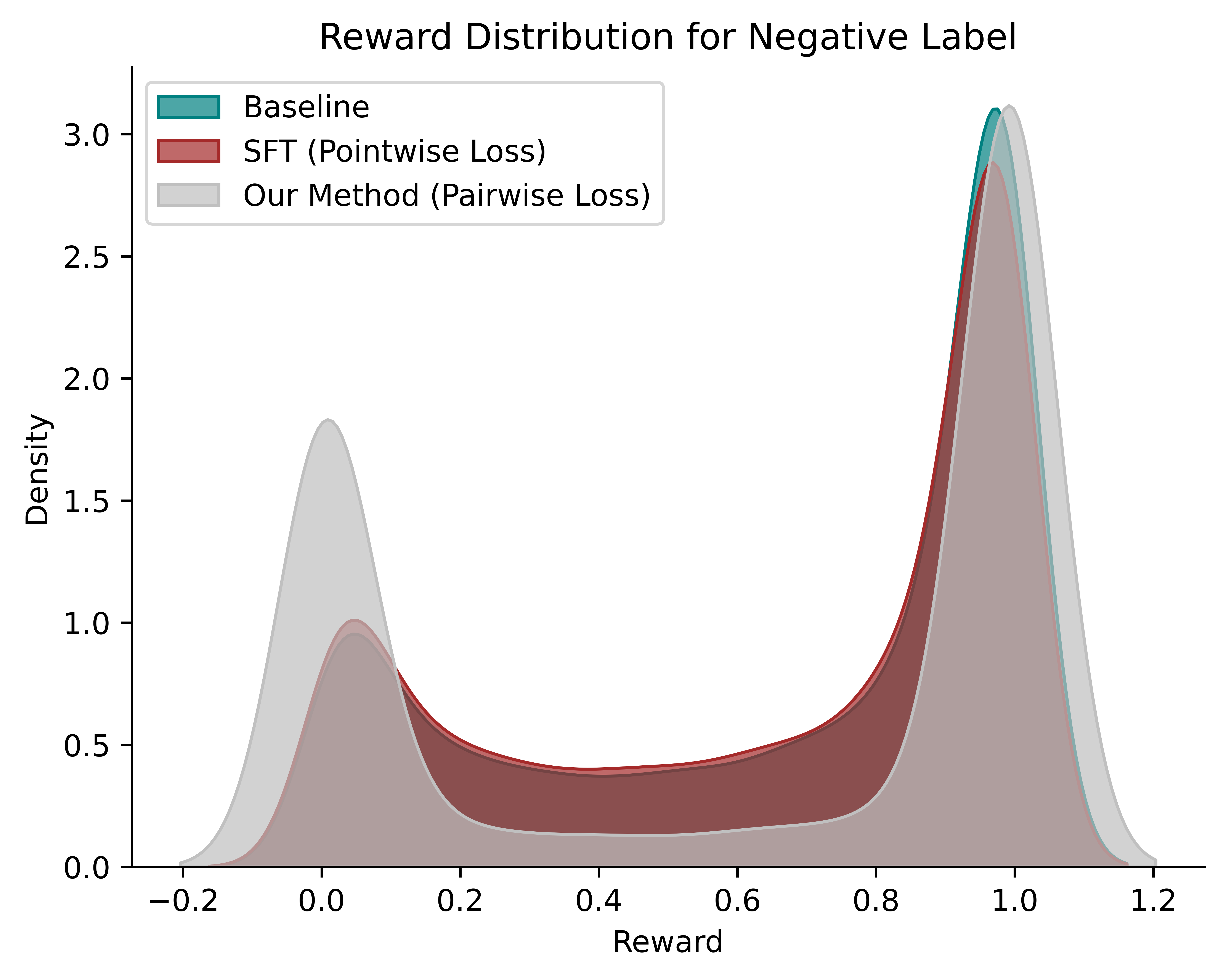}
    \caption{Reward distributions for negative labels across different QwenPRM variants. Green denotes the baseline QwenPRM, red denotes the model trained with default cross-entropy loss, and silver denotes the model trained with pairwise loss. The pairwise-trained model places substantially more mass around lower rewards for negative steps than the default PRM training method, indicating stronger separation of incorrect steps.} 
    \label{fig:reward_dist_pair_point_appx}
\end{figure}

\textbf{Takeaway.}
These results show that pairwise training reduces false positives more effectively than pointwise loss, which is the default training objective. Figure~\ref{fig:reward_dist_pair_point_appx} additionally compares the reward distributions induced by pointwise and pairwise losses for PRM training.

\begin{table}[t]
\centering
\caption{False-positive rate (FPR) comparison for pointwise- and pairwise-trained models.}
\label{tab:fpr_comparison}
\begin{tabular}{lcc}
\toprule
\textbf{Model} & \textbf{False Positives} & \textbf{False-Positive Rate (\%)} \\
\midrule
Baseline & 9109 & 69.34 \\
SFT (Pointwise Loss) & 8836 & 67.26 \\
Pairwise Loss & \textbf{8112} & \textbf{61.75} \\
\bottomrule
\end{tabular}
\end{table}

\subsubsection{Main Claim: Pairwise Training Reduces FPs at Fixed TPR.}\label{appx:TRP_FPR_theorem}

\textbf{Assumption}: The analysis focuses on the practically relevant regime where $N$ is moderate to large and the marginal predicted-positive rate $q=\mathbb{P}(\hat{T}=1)$ is not vanishing. In this setting, BoN selects from the predicted-positive set with high probability, and the dominant contribution to $P^{(N)}$ arises from selection among predicted positives; the contribution from the event where no predicted positive is present becomes negligible.

\begin{theorem}[ROC dominance $\implies$ lower FPR]
\textbf{Assumption:} Suppose pairwise training produces scores whose positive distribution dominates the negative distribution more than CE does. This assumption has been validated from distribution plots of Figure \ref{fig:reward_dist_pair_point_appx}. Concretely, assume there exists $\delta>0$ such that
\[
r_{\theta_{\text{pair}}}(y^{pos}) \overset{d}{=} r_{\theta_{\text{ce}}}(y^{pos}) + \delta,
\qquad
r_{\theta_{\text{pair}}}(y^{neg}) \overset{d}{=} r_{\theta_{\text{ce}}}(y^{neg}) - \delta.
\]

where $r_{\theta_{\text{pair}}}(y)$ is reward, $r_{\theta} (x, y_{<t}, y_t)$ from the PRM trained using pairwise loss and $r_{\theta_{\text{ce}}}(y)$ is the reward from the PRM trained using pointwise loss (cross-entropy loss).

Then for every recall/TPR level $\rho\in(0,1)$, the FPR of the Pairwise-trained model is no larger:
\[
\mathrm{FPR}_{\theta_{\text{pair}}}(\tau_{\text{pair}}(\rho)) 
\;\le\; \mathrm{FPR}_{\theta_{\text{ce}}}(\tau_{\text{ce}}(\rho)),
\]
where $\tau_{\text{pair}}(\rho)$ and $\tau_{\text{ce}}(\rho)$ are thresholds chosen to achieve $\mathrm{TPR}=\rho$ under each model.
\end{theorem}

\paragraph{Proof Sketch.}
Under the additive shift assumption, for any $\tau$,
\[
\Pr[r_{\theta_{\text{pair}}}(y^{pos}) \ge \tau] 
= \Pr[r_{\theta_{\text{ce}}}(y^{pos}) \ge \tau - \delta], 
\]
\[
\Pr[r_{\theta_{\text{pair}}}(y^{neg}) \ge \tau] 
= \Pr[r_{\theta_{\text{ce}}}(y^{neg}) \ge \tau + \delta].
\]

Additive shift assumption is valid after looking at the distribution plot of Figure \ref{fig:reward_dist_pair_point_appx}.

To attain the same TPR $\rho$, the BT model uses a higher threshold: 
$\tau_{\text{pair}}(\rho) = \tau_{\text{ce}}(\rho) + \delta$. Plugging into FPR:
\begin{align}
\mathrm{FPR}_{\theta_{\text{pair}}}(\tau_{\text{pair}}(\rho))
&= \Pr[r_{\theta_{\text{pair}}}(y^{neg}) \ge \tau_{\text{pair}}(\rho)] \\
&= \Pr[r_{\theta_{\text{ce}}}(y^{neg}) \ge \tau_{\text{pair}}(\rho) + \delta] \\
&= \Pr[r_{\theta_{\text{ce}}}(y^{neg}) \ge \tau_{\text{ce}}(\rho) + 2*\delta] \\
&\le \Pr[r_{\theta_{\text{ce}}}(y^{neg}) \ge \tau_{\text{ce}}(\rho)]
= \mathrm{FPR}_{\theta_{\text{ce}}}(\tau_{\text{ce}}(\rho)).
\end{align}
Thus, at any fixed recall/TPR, the BT model yields weakly smaller FPR. 

\paragraph{Interpretation.}
The pairwise objective enlarges the score margin 
$\Delta = s(X^+)-s(X^-)$, which uniformly lifts the ROC curve. ROC dominance implies lower FPR for any desired TPR, i.e., fewer false positives. 


\newpage

\section{Experimental Details and Additional Results}

\subsection{Samples per Curriculum Stage} \label{appx:CL_samplenum}

\begin{table}[htbp]
\centering
\resizebox{0.5\columnwidth}{!}{
\begin{tabular}{lc}
\hline
\textbf{} & \textbf{Number of Samples.}  \\
\midrule
\multicolumn{2}{c}{\emph{Qwen-PRM-7B trained on 12k paired data.}} \\
\midrule
CL1 (0.5-1) & 7.0k  \\
CL2 (0.3-0.5) & 2.0k  \\
CL3 (0.2-0.3) & 2.0k  \\
CL4 (0.1-0.2) & 1.0k  \\
\midrule
\multicolumn{2}{c}{\emph{Qwen-PRM-7B trained on 132k paired data.}} \\
\midrule
CL1 (0.5-1) & 75.0k  \\
CL2 (0.3-0.5) & 23.5k  \\
CL3 (0.2-0.3) & 23.5k  \\
CL4 (0.1-0.2) & 10.0k  \\
\midrule
\end{tabular}
}
\caption{Number of samples in each curriculum learning round.}
\end{table}


\subsection{ProcessBench Results}\label{appx:processbench}

ProcessBench\citep{zheng2025processbench} evaluation shows the same overall pattern observed on PRMBench: lowering FPR is accompanied by a decrease in TPR.

\begin{table}[hbt!]
\centering
\resizebox{\textwidth}{!}{%
\begin{tabular}{lcccccccccccc}
\toprule
 & \multicolumn{3}{c}{\textbf{GSM8K}} & \multicolumn{3}{c}{\textbf{MATH500}} & \multicolumn{3}{c}{\textbf{Omnimath}} & \multicolumn{3}{c}{\textbf{OlympiadBench}} \\
\cmidrule(lr){2-4} \cmidrule(lr){5-7} \cmidrule(lr){8-10} \cmidrule(lr){11-13}
 & \textbf{TPR} & \textbf{TNR} & \textbf{F1} & \textbf{TPR} & \textbf{TNR} & \textbf{F1} & \textbf{TPR} & \textbf{TNR} & \textbf{F1} & \textbf{TPR} & \textbf{TNR} & \textbf{F1} \\
\midrule
Baseline & 71.0 & 96.4 & 81.8 & 67.0 & 91.1 & 77.2 & 54.8 & 66.1 & 83.4 & 54.6 & 85.5 & 66.7 \\
PRISM CL1 & 72.0 & 94.8 & 81.8 & 69.4 & 86.9 & 77.2 & 55.6 & 66.2 & 81.7 & 58.2 & 79.6 & 67.3 \\
PRISM CL2 & 73.4 & 94.3 & 82.6 & 70.5 & 85.0 & 77.1 & 58.2 & 65.8 & 75.5 & 58.9 & 75.5 & 66.1 \\
PRISM CL3 & 73.4 & 93.8 & 82.4 & 72.6 & 82.0 & 77.0 & 57.8 & 64.8 & 72.6 & 58.5 & 72.6 & 64.8 \\
\bottomrule
\end{tabular}%
}
\caption{}
\end{table}

\subsection{Detailed PRMBench Results}\label{appx:prmbench_results}

\begin{table}[hbt!]
\centering
\resizebox{1\columnwidth}{!}{
\begin{tabular}{l c ccc ccccc cccc}
\toprule
\multirow{2}{*}{\textbf{Model}} & \multirow{2}{*}{\textbf{Overall}} &
\multicolumn{3}{c}{\textbf{Simplicity}} &
\multicolumn{5}{c}{\textbf{Soundness}} &
\multicolumn{4}{c}{\textbf{Sensitivity}} \\
\cmidrule(lr){3-5}\cmidrule(lr){6-10}\cmidrule(lr){11-14}
 &  & \textbf{NR.} & \textbf{NCL.} & \textbf{Avg.} &
 \textbf{ES} & \textbf{SC.} & \textbf{DC.} & \textbf{CI} & \textbf{Avg.} &
 \textbf{PS} & \textbf{DR.} & \textbf{MS.} & \textbf{Avg.} \\
 \midrule
\multicolumn{14}{c}{\emph{Qwen-PRM-7B}} \\
\midrule
Baseline & 65.5 & 49.1 & 55.0 & 52.1 & 71.7 & 67.4 & 66.3 & \textbf{78.5} & 71.0 & 57.7 & 69.1 & \textbf{99.7} & 75.5 \\
\midrule
\multicolumn{14}{c}{\emph{Qwen-PRM-7B trained on 12k paired data.}} \\
\midrule
Without Curriculum & 67.0 & 50.6 & 59.3 & 54.9 & 73.2 & 68.4 & 66.9 & 78.0 & 71.6 & 59.0 & 70.7 & 99.6 & 76.5 \\
\midrule
Curriculum-1 & 66.7 & 50.5 & 58.6 & 54.5 & 72.8 & {68.1} & 67.3 & {77.8} & {71.5} & 58.8 & 70.4 & 99.6 & 76.3 \\
Curriculum-2 & 67.2 & 51.4 & 60.2 & 55.8 & {73.3} & 68.0 & {67.7} & 76.4 & 71.4 & 60.4 & {70.4} & 99.3 & 76.7 \\
Curriculum-3 & {67.3} & {52.5} & {63.2} & {57.8} & 73.2 & 67.2 & 67.5 & 75.0 & 70.7 & {61.9} & 69.8 & 98.9 & {76.9} \\
\midrule
\multicolumn{14}{c}{\emph{Qwen-PRM-7B trained on 132k paired data.}} \\
\midrule
Curriculum-1 & 67.0 & 50.4 & 59.3 & 54.9 & 72.9 & 68.9 & 67.5 & 78.4 & 72.0 & 58.8 & 70.5 & 99.7 & 76.3\\ 
Curriculum-2 & 67.8 & 51.4 & 61.9 & 56.6 & 73.9 & 69.1 & 67.7 & 78.1 & 72.2 & 60.0 & 70.8 & 99.6 & 76.8\\ 
Curriculum-3 & 68.0 & 52.3 & 63.7 & 58.0 & \underline{73.9} & 68.5 & 67.4 & 77.4 & 71.8 & 61.1 & 70.6 & 99.3 & 77.0\\ 
Curriculum-4 & 68.0 & 53.5 & \textbf{66.0} & \textbf{59.7} & 73.8 & 67.5 & 67.1 & 75.6 & 71.0 & 62.4 & 70.1 & 98.7 & \underline{77.1}\\
\midrule
\end{tabular}
}
\caption{PRMBench results for Qwen2.5-Math-PRM-7B (baseline) and its trained variants across different metrics. Columns are grouped into \emph{Simplicity}, \emph{Soundness}, and \emph{Sensitivity}. Bold indicates the best score within each sub-metric, with PRMScore reported as the final comparison metric.}
\label{tab:PRMbench_results}
\end{table}

\subsection{ReasonEval-7B Results}\label{appx:reasoneval_results}

Results on another PRM model, ReasonEval-7B, are provided below to assess generalizability. Similar behavior appears here as well: later curriculum rounds substantially reduce FPR with only a slight increase in FNR, while PRMScore improves steadily. This model also appears in Table~\ref{tab:intro_SOTA_PRMs}.

\begin{table}[!htbp]
\centering
\begin{tabular}{lccc}
\toprule
\textbf{Threshold / Curriculum} & \textbf{PRM Score} & \textbf{FPR} & \textbf{FNR} \\
\midrule
CL1 (0.5 -- 1.0) & 61.0 & 78.79 & 4.21 \\
CL2 (0.3 -- 0.5) & 61.8 & 69.76 & 8.05 \\
CL3 (0.1 -- 0.3) & 62.3 & 66.8  & 9.9  \\
CL4 (0.0 -- 0.1) & 62.3 & 61.7  & 13.6 \\
\bottomrule
\end{tabular}
\caption{Performance of ReasonEval-7B PRM under the PRISM training recipe, illustrating generalization beyond Qwen-PRM.}
\label{tab:curriculum_prm}
\end{table}

\subsection{Curriculum Bin Ablations}\label{appx:cl_bins_ablation}

Multiple threshold settings for curriculum learning are reported in Tables~\ref{tab:exp1} and \ref{tab:exp2}. Final performance remains consistent across different bin configurations, which is expected given the gradual progression toward more confusing or difficult regions. Across all curriculum-learning bin configurations, the false-positive rate decreases substantially while the false-negative rate increases only slightly as training advances through the curriculum rounds.

\begin{table}[!htbp]
\centering
\begin{tabular}{lccc}
\toprule
\textbf{Threshold / Curriculum} & \textbf{PRM Score} & \textbf{FPR} & \textbf{FNR} \\
\midrule
CL1 (0.7 -- 1.0) & 66.4 & 65.58 & 5.4 \\
CL2 (0.5 -- 0.7) & 67.2 & 62.4  & 6.29 \\
CL3 (0.3 -- 0.5) & 68.0 & 56.6  & 8.24 \\
CL4 (0.1 -- 0.3) & 68.2 & 49.0  & 11.0 \\
\bottomrule
\end{tabular}
\caption{Results on Qwen-PRM-7B training under the PRISM recipe with one curriculum-bin configuration.}
\label{tab:exp1}
\end{table}

\begin{table}[!htbp]
\centering
\begin{tabular}{lccc}
\toprule
\textbf{Threshold / Curriculum} & \textbf{PRM Score} & \textbf{FPR} & \textbf{FNR} \\
\midrule
CL1 (0.5 -- 1.0) & 67.4 & 61.25  & 6.65 \\
CL2 (0.3 -- 0.5) & 68 & 54.61 & 9.16  \\
CL3 (0.1 -- 0.3) & 67.8 & 45.70  & 13.56  \\
\bottomrule
\end{tabular}
\caption{Results on Qwen-PRM-7B training under the PRISM recipe with an alternative curriculum-bin configuration.}
\label{tab:exp2}
\end{table}

\subsection{Answer Selection Methods}\label{appx:answer_selection}
\paragraph{Setup:} Consider $N$ candidate solutions sampled from a model for the same problem.  
Each solution consists of:
\begin{enumerate}
    \item A \textbf{final answer}, e.g., a number in GSM8K or an option in ARC.
    \item A \textbf{verifier score}, assigned by a process reward model or external verifier, indicating how plausible or correct the reasoning chain appears.
\end{enumerate}

\paragraph{Simple Majority Voting:} In plain majority voting, completions are grouped by their \textbf{final answer}.
\begin{itemize}
    \item Count how many completions lead to each distinct answer.
    \item Select the answer with the largest count.
\end{itemize}
Formally, if $c(a)$ is the number of completions yielding answer $a$, then the majority-vote answer is:
\[
a^* = \arg\max_a c(a).
\]

\paragraph{Verifier-Weighted Majority Voting:} Instead of giving each completion equal weight, votes are weighted by verifier scores.  
Let answer $a$ appear in solutions $\{s_1, s_2, \ldots, s_k\}$, where each solution $s_i$ has verifier score $v(s_i)$.  
The total verifier-weighted score for answer $a$ is:
\[
V(a) = \sum_{s_i : \,\text{final}(s_i) = a} v(s_i).
\]
The answer with the largest weighted score is then selected:
\[
a^* = \arg\max_a V(a).
\]
This approach discounts low-scoring (less credible) reasoning chains and prefers answers supported by higher-quality solutions.

\subsection{OOD Guided Beam Search Results}

\begin{figure}[H]
    \centering
    \includegraphics[width=0.5\linewidth]{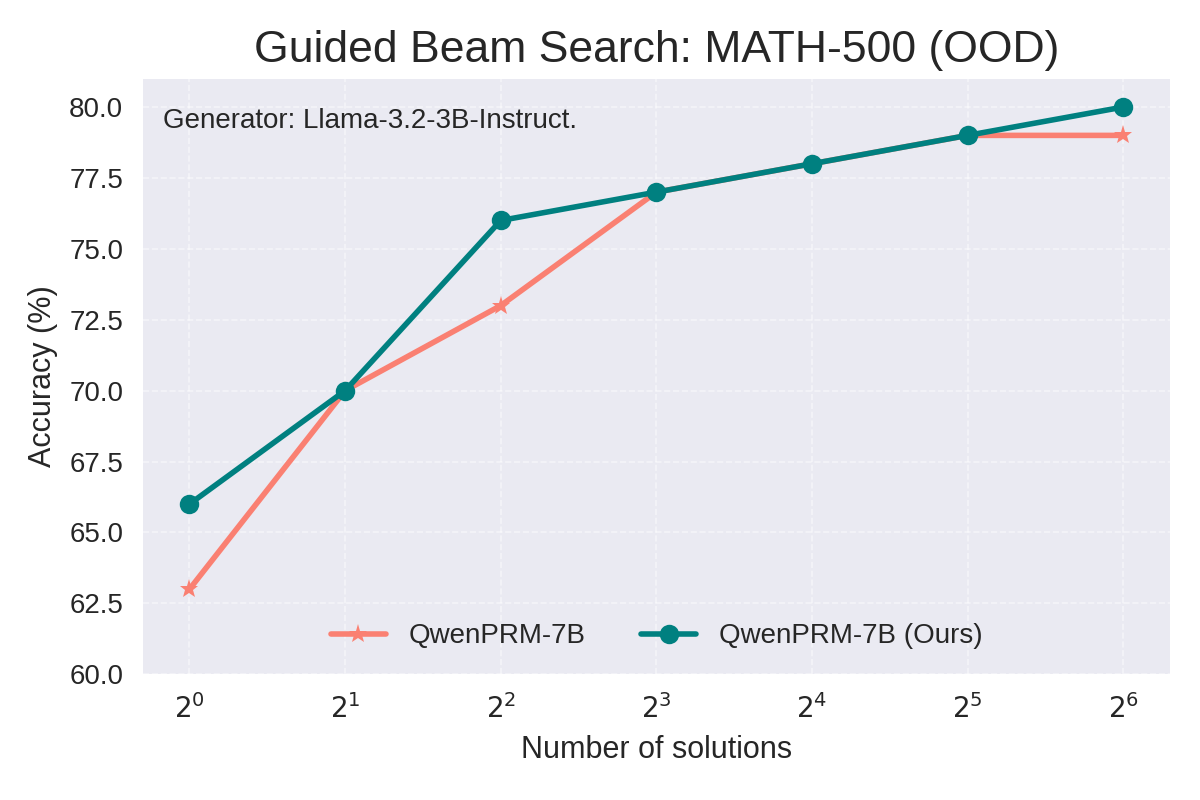}
    \caption{OOD guided beam-search results on MATH-500. PRISM (blue) substantially outperforms the baseline verifier (orange), even when the generator policy is out of distribution.}
    \label{fig:guided_OOD}
\end{figure}

\subsection{Statistical Plots}

\begin{figure}[H]
    \centering
    \includegraphics[width=\linewidth, trim={0 0 0 0}, clip]{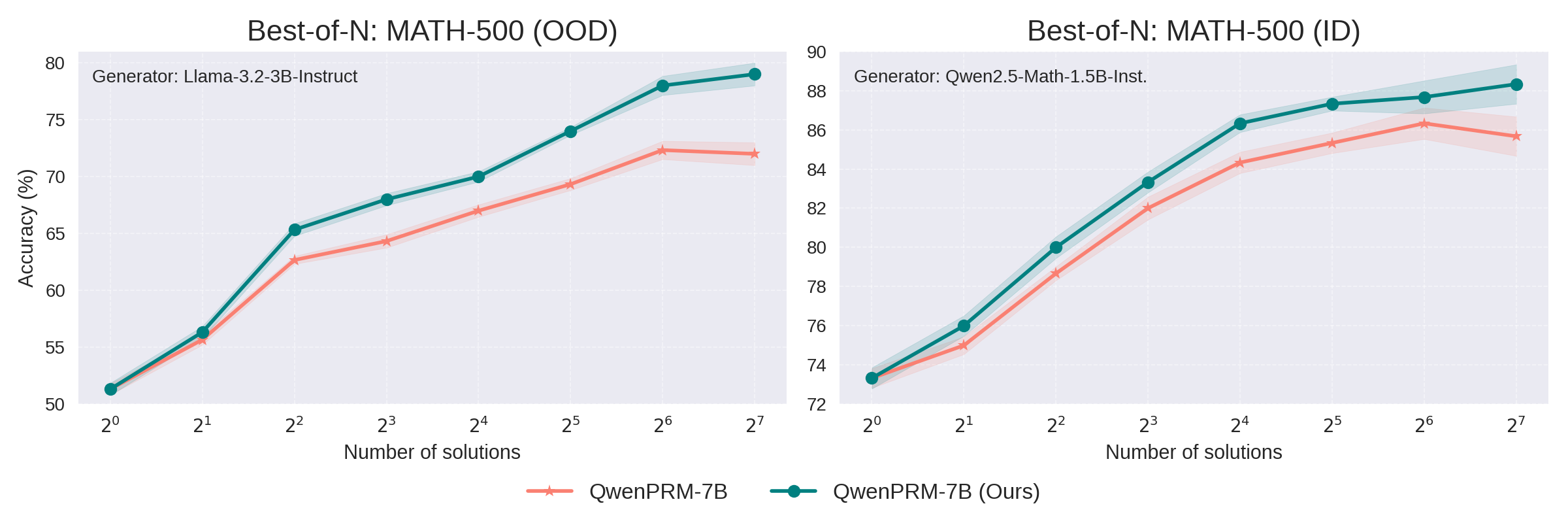}
    \vspace{-2em}
    \caption{Error plot comparing best-of-$N$ alignment on MATH-500 using two generator policies: LLaMA for the OOD policy (left) and Qwen for the ID policy (right). The baseline PRM is shown in orange and PRISM in blue. Across both settings, PRISM delivers a clear performance boost over the baseline, with the improvement being especially pronounced for the OOD policy.}
    \label{fig:best_of_N_ID_OOD_error}
    \vspace{-3mm}
\end{figure}

